\documentclass[10pt,journal,compsoc]{IEEEtran}
\usepackage[hidelinks]{hyperref}
\usepackage{url}            
\usepackage{booktabs}       
\usepackage{amsfonts}       
\usepackage{nicefrac}       
\usepackage{microtype}      
\usepackage{xcolor}         
\usepackage{graphicx}
\usepackage{amsmath}
\usepackage{makecell}
\usepackage{multirow}
\usepackage{algorithm}
\usepackage{algpseudocode}
\usepackage{color}
\usepackage{xcolor}
\usepackage{etoolbox}

\usepackage{caption}
\usepackage{colortbl}
\ifCLASSOPTIONcompsoc
  \usepackage[nocompress]{cite}
\else
  \usepackage{cite}
\fi

\ifCLASSINFOpdf
\else
\fi

\newcommand{\myparagraph}[1]{\vspace{0.1em}\noindent\textbf{#1}}
\definecolor{mygreen}{rgb}{0, 0, 0}
\definecolor{myred}{rgb}{0, 0, 0}

\usepackage{ulem}

\newcommand{\whatsnew}[1]{{#1}}
\newcommand{\suspicious}[1]{{#1}}

\makeatletter
\patchcmd{\@makecaption}
  {\scshape}
  {}
  {}
  {}
\makeatother

\begin{document}

\title{Vote2Cap-DETR++: Decoupling Localization and Describing for End-to-End 3D Dense Captioning}

\author{
    Sijin Chen$^{1*}$,
    Hongyuan Zhu$^{2}$, ~\IEEEmembership{Member,~IEEE},
    Mingsheng Li$^{1}$,
    Xin Chen$^{3}$,
    Peng Guo$^{1}$,
    Yinjie Lei$^{4}$, ~\IEEEmembership{Member,~IEEE},
    Gang YU$^{3}$, ~\IEEEmembership{Member,~IEEE},
    Taihao Li$^{5}$, 
    Tao Chen$^{1\dagger}$, ~\IEEEmembership{Senior Member,~IEEE}
    \\
    \thanks{
        $^*$This work was partially accomplished under the supervision by Dr. Hongyuan Zhu from A*STAR, Singapore.
    }
    \thanks{$^{\dagger}$Corresponding author.} 
    \thanks{
        $^{1}$Sijin Chen, Mingsheng Li, Peng Guo, and Tao Chen are with the School of Information Science and Technology, Fudan University, Shanghai 200433, China (Corresponding author: Tao Chen, e-mail: \href{eetchen@fudan.edu.cn}{eetchen@fudan.edu.cn}, tel: +86-2131242503).
    }
    \thanks{
        $^{2}$Hongyuan Zhu is with the Institute for Infocomm Research (I$^2$R) \& Centre for Frontier AI Research (CFAR), A*STAR, Singapore.
    }
    \thanks{$^{3}$Xin Chen and Gang Yu are with Tencent PCG, China.}
    \thanks{$^{4}$Yinjie Lei is with Sichuan University.}
    \thanks{
        $^{5}$Taihao Li is with Zhejiang Lab, Hangzhou, China.
    }
}



\IEEEtitleabstractindextext{
    \begin{abstract}
3D dense captioning requires a model to translate its understanding of an input 3D scene into several captions associated with different object regions.
Existing methods adopt a sophisticated “detect-then-describe” pipeline, which builds explicit relation modules upon a 3D detector with numerous hand-crafted components.
While these methods have achieved initial success, the cascade pipeline tends to accumulate errors because of duplicated and inaccurate box estimations and messy 3D scenes.
In this paper, we first propose Vote2Cap-DETR, a simple-yet-effective transformer framework that decouples the decoding process of caption generation and object localization through parallel decoding.
%
%
\whatsnew{
Moreover, we argue that object localization and description generation require different levels of scene understanding, which could be challenging for a shared set of queries to capture.
To this end, we propose an advanced version, Vote2Cap-DETR++, which decouples the queries into localization and caption queries to capture task-specific features.
Additionally, we introduce the iterative spatial refinement strategy to vote queries for faster convergence and better localization performance.
We also insert additional spatial information to the caption head for more accurate descriptions.
Without bells and whistles, extensive experiments on two commonly used datasets, ScanRefer and Nr3D, demonstrate Vote2Cap-DETR and Vote2Cap-DETR++ surpass conventional ``detect-then-describe'' methods by a large margin.
}
Codes will be made available at \href{https://github.com/ch3cook-fdu/Vote2Cap-DETR}{https://github.com/ch3cook-fdu/Vote2Cap-DETR}.
\end{abstract}


}

\maketitle

\section{Introduction}
\label{sec:intro}

\begin{figure}[htbp]
	\centering
	\includegraphics[width=\linewidth]{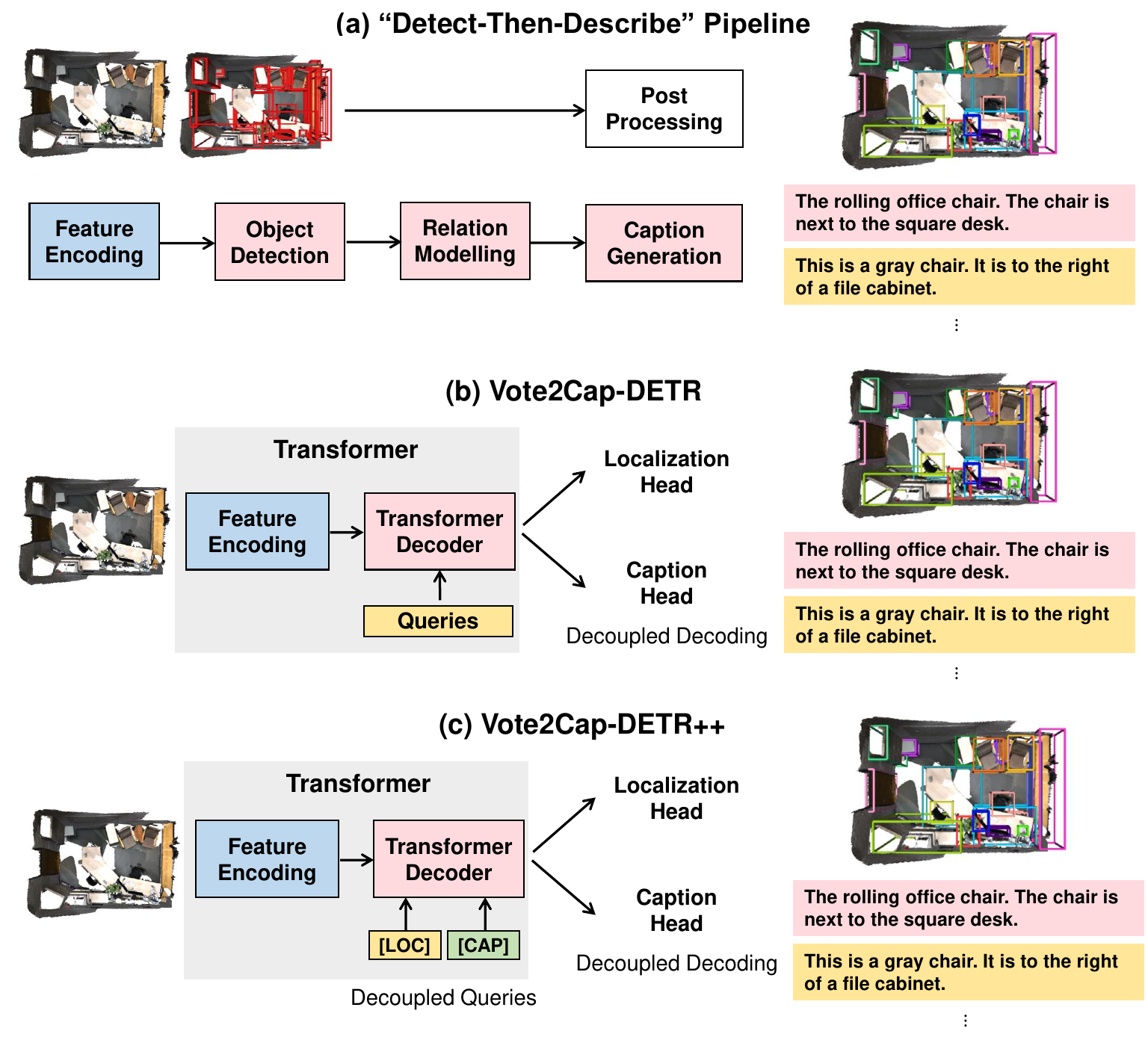}
    \vspace{-5mm}
	\caption{
    	\textbf{A comparison between our proposed methods and conventional ``detect-then-describe'' pipelines.}
        (a) The ``detect-then-describe'' pipelines build explicit relation modules directly on a detector's box-estimations, resulting in cumulative errors and heavy reliance on hand-crafted components. 
        (b) The proposed Vote2Cap-DETR frames 3D dense captioning as a set prediction problem and decouple the decoding process of object localization and caption generation. 
        (c) In the proposed Vote2Cap-DETR++, we further decouple the queries to capture task-specific features.
    }
	\label{fig:teaser}
\end{figure}

\IEEEPARstart{R}{ecent} years have witnessed significant growth in the field of 3D learning for various applications\cite{guo2020deep,xiao2023unsupervised,chen2023executing,yin2022coordinates,lin2021learning,hu2021learning,meng2021towards}. 
As part of this trend, the task of 3D dense captioning\cite{chen2021scan2cap} has emerged, which requires a model to localize and generate descriptive sentences for all objects within an input 3D scene. 
This problem is challenging, given 1) the sparsity of a point cloud and 2) the cluttered 3D scene.

Prior works have achieved great success in 3D dense captioning.
Scan2Cap\cite{chen2021scan2cap}, SpaCap3D\cite{wang2022spacap3d}, MORE\cite{jiao2022more}, and REMAN\cite{mao2023complete} extract relations among box estimations with well-designed relation modeling modules.
Concurrently, \cite{zhong2022contextual3DdenseCap} introduces an additional contextual branch to capture non-object information.
3DJCG\cite{cai20223djcg}, D3Net\cite{chen2021d3net}, and 3D-VLP\cite{jin2023context} study the mutual promotion of various \textbf{3D} \textbf{V}ision-\textbf{L}anguage (3DVL) tasks, containing additional tasks like \textbf{3D} \textbf{V}isual \textbf{G}rounding (3DVG), \textbf{3D} \textbf{Q}uestion \textbf{A}nswering (3DQA).
$\chi$-Trans2Cap\cite{yuan2022x-trans2cap} also shows that transferring knowledge from additional 2D information could also improve the quality of captions generated.

Among the existing methods, they all adopt the ``detect-then-describe'' pipeline (Figure \ref{fig:teaser}, upper).
Specifically, they perform object localization and description generation in a cascade way by modeling relations among box estimations.
Though these methods have achieved remarkable performance, the ``detect-then-describe'' pipeline suffers from the following issues: 
1) Because of the serial and explicit reasoning, the latter modules depend heavily on object detection performance.
The duplicate box predictions cause confusion and limit the mutual promotion of detection and captioning.
2) The pipeline requires a vast number of hand-crafted components, such as 3D operators\cite{qi2017pointnet++}, relation graphs within box estimations\cite{chen2021scan2cap,jiao2022more}, and \textbf{N}on-\textbf{M}aximum \textbf{S}uppression (NMS)\cite{neubeck2006nms} for post-processing.
These hand-crafted components introduce additional hyper-parameters, leading to a sub-optimal performance given the sparse object surfaces and messy indoor scenes.
%

To address the above issues, we first propose a preliminary model named Vote2Cap-DETR, which is a full transformer\cite{vaswani2017attention} encoder-decoder model that decouples decoding process of caption generation and object localization in 3D dense captioning.
Unlike the conventional ``detect-then-describe'' pipeline, Vote2Cap-DETR decouples the captioning process from object localization by applying two parallel task heads.
By further casting 3D dense captioning to a set-to-set problem, we associate each target instance and its language annotation with a distinctive query, encouraging the models to learn more discriminative proposal representations, which in turn helps to identify each distinctive object in a 3D scene.
To further facilitate the model's localization ability, we propose a novel vote decoder by reformulating the object queries in 3DETR\cite{misra2021-3detr} into the format of vote queries, which is a composition of the embedding and the vote spatial transformation of seed points.
This also builds the connection between the vote queries in Vote2Cap-DETR and the VoteNet\cite{qi2019votenet}, but with a higher localization capacity.
%
Besides, we develop a novel query-driven caption head that captures relation and attribute information through self- and cross-attention for descriptive and object-centric object captions.

\whatsnew{
Though Vote2Cap-DETR has established an elegant decoupled decoding approach to 3D dense captioning, it still has certain limitations.
Localizing scene objects necessitates different levels of scene understanding from generating informative scene object descriptions.
The former relies on a model's perception of an object's 3D structures to generate tight bounding boxes, while the latter relies on sufficient attribute information and spatial relations.
Therefore, decoding the same set of queries to descriptions and box estimations makes it difficult for a model to capture discriminative features for either task, resulting in a sub-optimal performance.
}

\whatsnew{
To address this issue, we further introduce an advanced framework, namely Vote2Cap-DETR++, to remove the obstacles to extracting task-specific features.
As is shown in Figure \ref{fig:teaser} (bottom), we decouple the queries into 3D localization queries (``[LOC]'') and caption queries (``[CAP]'') with a shared transformer decoder for decoupled sub-task decoding.
The two sets of queries correspond with each other, as they are tied to the same box-caption estimation.
We further propose two additional strategies for better object localization and caption generation.
First, we introduce the iterative refinement strategy for vote queries in the transformer decoder to progressively shorten the distance between query points and objects.
This leads to faster convergence and better detection performance. 
In addition, we insert an additional 3D positional encoding token to the caption prefix and apply a rank-based positional encoding for local surrounding guidance to help the caption head identify the exact location of a query for accurate caption generation.
We empirically show that the advanced model, Vote2Cap-DETR++, performs consistently better than the preliminary version through extensive experiments.
}

\whatsnew{
The preliminary version is published in \cite{chen2023vote2cap}.
Compared to that, we have made significant improvements and extensions in three folds.
We propose the decoupled-and-correspond queries to capture task-specific features for object localization and caption generation.
Besides, we introduce a spatial refinement strategy on vote queries for faster convergence and better detection performance.
Concurrently, we insert 3D spatial information into the caption generation process for more accurate descriptions.
To the best of our knowledge, this is the first non-``detect-then-describe'' method for 3D dense captioning.
With extensive experiments, we show that the advanced framework outperforms the preliminary version by a large margin.
To facilitate and inspire further research in 3D dense captioning, we have made our codes partially available in \href{https://github.com/ch3cook-fdu/Vote2Cap-DETR}{https://github.com/ch3cook-fdu/Vote2Cap-DETR}, and we will keep updating codes for the advanced version.
}

Experiments on two commonly used datasets demonstrate that both proposed approaches surpass prior ``detect-then-describe'' approaches with many hand-crafted components by a large margin.
Our preliminary framework, Vote2Cap-DETR, achieves 73.77\% and 45.53\% C@0.5 on the validation set of ScanRefer\cite{chen2020scanrefer} and Nr3D\cite{achlioptas2020referit3d}, respectively.
Remarkably, the advanced version, Vote2Cap-DETR++, further achieves 78.16\% C@0.5 (\textcolor{mygreen}{+4.39}\%) and 47.62\% C@0.5 (\textcolor{mygreen}{+2.09}\%), which surpasses the preliminary version, and sets new state-of-the-art records on both datasets.

To summarize, this paper's main contributions include:

\begin{itemize} 
\setlength\itemsep{0em}
    \item We propose two transformer-based 3D dense captioning frameworks that decouple the caption generation from object localization to avoid the cumulative errors brought up by the explicit reasoning on box estimations in ``detect-then-describe'' pipelines.

    \item We decouple the decoding process and feature extraction in 3D dense captioning to help the model learn discriminative features for object localization and description generation. 
    By further introducing the iterative spatial refinement strategy for queries and incorporating additional spatial information into caption generation, our method can generate bounding boxes and descriptions in a higher quality.
    
    \item Extensive experiments show that both the proposed Vote2Cap-DETR and Vote2Cap-DETR++ set new state-of-the-arts on both Nr3D\cite{achlioptas2020referit3d} and ScanRefer\cite{chen2021scan2cap}.
\end{itemize}

\whatsnew{
The remainder of this paper is organized as follows.
We first briefly introduce the related works on 3D Vision-Language tasks, 3D dense captioning, DETRs, and other visual captioning tasks in Section \ref{sec:related}.
Then, we deliver basic information for transformers before introducing our proposed Vote2Cap-DETR and Vote2Cap-DETR++ in Section \ref{sec:method}.
After that, we take out extensive experiments and visualizations to verify the effectiveness of our proposed methods in Section \ref{sec:exp}.
Finally, we state the limitations of our work in Section \ref{sec:limitations} and draw the conclusions in Section \ref{sec:conclusion}.
}
\section{Related Works}
\label{sec:related}

In this section, we first cover existing trends in 3D vision-language tasks, and summarize existing approaches in 3D dense captioning.
After that, we show recent advances on transformers for both 2D and 3D object detection.
We also include related works on other caption tasks.

\subsection{3D Vision-Language Tasks}

\whatsnew{
\textbf{3D} \textbf{V}ision-\textbf{L}anguage (3DVL) tasks require a model to showcase its understanding of a 3D scene in response to, or in response by language.
\textbf{3D} \textbf{D}ense \textbf{C}aptioning (3DDC)\cite{chen2021scan2cap,chen2023vote2cap} feeds a model with a 3D scene and expects a set of instance location estimations paired with natural language descriptions in response.
\textbf{3D} \textbf{V}isual \textbf{G}rounding (3DVG)\cite{chen2020scanrefer,achlioptas2020referit3d,zhao20213dvg,wu2023eda} requires a model to localize the one and only object mentioned in the query sentence from a 3D scene.
\textbf{3D} \textbf{Q}uestion \textbf{A}nswering (3DQA)\cite{ye20223dqa,azuma2022scanqa,ma2022sqa3d} expects a model to answer the question based on the corresponding 3D scene input.
The majority of the existing 3DVL approaches\cite{chen2020scanrefer,achlioptas2020referit3d,chen2021scan2cap,jin2023context,azuma2022scanqa,ye20223dqa} adopt the ``detect-then-describe'' pipeline that builds multi-modal fusion or spatial relation reasoning modules on top of a 3D detector\cite{qi2019votenet}, instance segmentation model\cite{jiang2020pointgroup}, or even ground truth instance labels\cite{achlioptas2020referit3d,he2021transrefer3d,roh2022languagerefer}.
In this paper, we propose two transformer-based 3DDC models that decouple the object localization and description generation process.
We hope this design will inspire follow-up works to rethink the design of models in 3DVL tasks.
}

\subsection{3D Dense Captioning}
3D dense captioning is a challenging task that requires a model to accurately localize and generate informative descriptions for all the objects from a cluttered 3D scene.
Since its inception, various methods have been proposed to tackle this challenging problem.
Scan2Cap\cite{chen2021scan2cap}, D3Net\cite{chen2021d3net}, REMAN\cite{mao2023complete} and MORE\cite{jiao2022more} treats each proposal of a 3D detector's box estimations as a graph node, and manually build k nearest neighbor graphs to extract features with graph operations\cite{wang2019edgeconv}.
3DJCG\cite{cai20223djcg}, SpaCap3D\cite{wang2022spacap3d}, 3D-VLP\cite{jin2023context}, $\chi$-Tran2Cap\cite{yuan2022x-trans2cap} and UniT3D\cite{chen2022unit3d} replace the graph operations with transformers\cite{vaswani2017attention} to capture spatial relations among object proposals.
Meanwhile, 3D-VLP\cite{jin2023context}, UniT3D\cite{chen2022unit3d}, D3Net\cite{chen2021d3net}, and 3DJCG\cite{cai20223djcg} shifts attention to the mutual promotion of various 3D vision-language tasks, including 3D dense captioning, 3D visual grounding\cite{chen2020scanrefer}, and even 3D question answering\cite{ye20223dqa,azuma2022scanqa}.
One can see that most of the above-mentioned methods all adopt the ``detect-then-describe'' pipeline, which directly builds spatial relation modules on the output of a 3D detector.
Though straightforward and simple as these approaches are, the explicit relation modeling procedure is sensitive to certain hyperparameters, such as the definition of edges between two nodes, and the number of nearest neighbors, and could be easily confused by the duplicated and inaccurate box estimations from the 3D detector.
Our proposed methods are able to bypass these limitations via decoupled and parallel sub-task decoding.
We further propose decoupled queries to capture task-specific features.

\subsection{DEtection TRansformers (DETR): from 2D to 3D}

DETRs\cite{carion2020detr,zhu2020deformabledetr} are transformer\cite{vaswani2017attention} based object detectors that treat the detection problem as set prediction, and are able to generate sparse predictions robust to \textbf{N}on-\textbf{M}aximum \textbf{S}uppression (NMS)\cite{neubeck2006nms} for post-processing.
Though astonishing performance has been achieved, the original DETR\cite{carion2020detr} suffers from slow convergence.
As a result, many follow-ups\cite{chen2022groupdetr,jia2022hybriddetrs,zhu2020deformabledetr,meng2021conditionaldetr} have made great attempts to accelerating the training procedure by exploring the label assignment strategy\cite{chen2022groupdetr,jia2022hybriddetrs,liu2023stabledetr}, the usage of multi-scale features\cite{zhang2022detr++,zhu2020deformabledetr,li2023litedetr}, and the design of attention\cite{chen2022conditionaldetrv2,zhu2020deformabledetr}.
Researchers also make great attempts to extend the idea of DETR to 3D object detection.
GroupFree3D\cite{liu2021groupfree3d} learns object proposals from the input point cloud with transformers rather than grouping local points\cite{qi2019votenet}, and 3DETR\cite{misra2021-3detr} explores the potential of the standard transformer architecture for 3D object detection.
In this paper, we further extend the idea of DETR to treat 3D dense captioning as a set prediction problem, which learns to predict a set of object-caption pairs directly from a set of points.

\subsection{Other Visual Captioning Tasks}

Image captioning and video dense captioning are also two important tasks in multi-modal learning.
An image captioning model is capable of understanding and describing the key elements/relations in an input image.
The current trend\cite{anderson2018bottom,huang2019AinA-img-cap,nguyen2022grit} in image captioning is to adopt an encoder-decoder architecture for generating sentences that describe the key elements of the entire image.
Among those approaches, \cite{anderson2018bottom,huang2019AinA-img-cap,cornia2020meshed} encode the input image with region features extracted by a pre-trained image detector\cite{ren2015faster}, while \cite{mokady2021clipcap,liu2021cptr} directly extract grid feature with an image encoder\cite{dosovitskiy2020image} pre-trained on ImageNet\cite{deng2009imagenet}.
More recently, \cite{luo2023semantic,zhu2022exploring} explore the potential of diffusion models\cite{ho2020denoising} in image captioning.
Concurrently, researchers also put much effort in video dense captioning.
Video dense captioning\cite{krishna2017dense} requires a model to segment the input video and describe the event in each video clip.
Recently, \cite{wang2021end,zhou2018end} adopt a transformer architecture for end-to-end video dense captioning.
In our paper, we focus on 3D dense caption, which translates the understanding of a 3D scene to boxes and words.
We also propose several key components designed especially for 3D tasks, including the designs for queries, and the local-context aware caption heads.

\section{Method}
\label{sec:method}

\begin{figure*}[htbp]
	\centering
	\includegraphics[width=\linewidth]{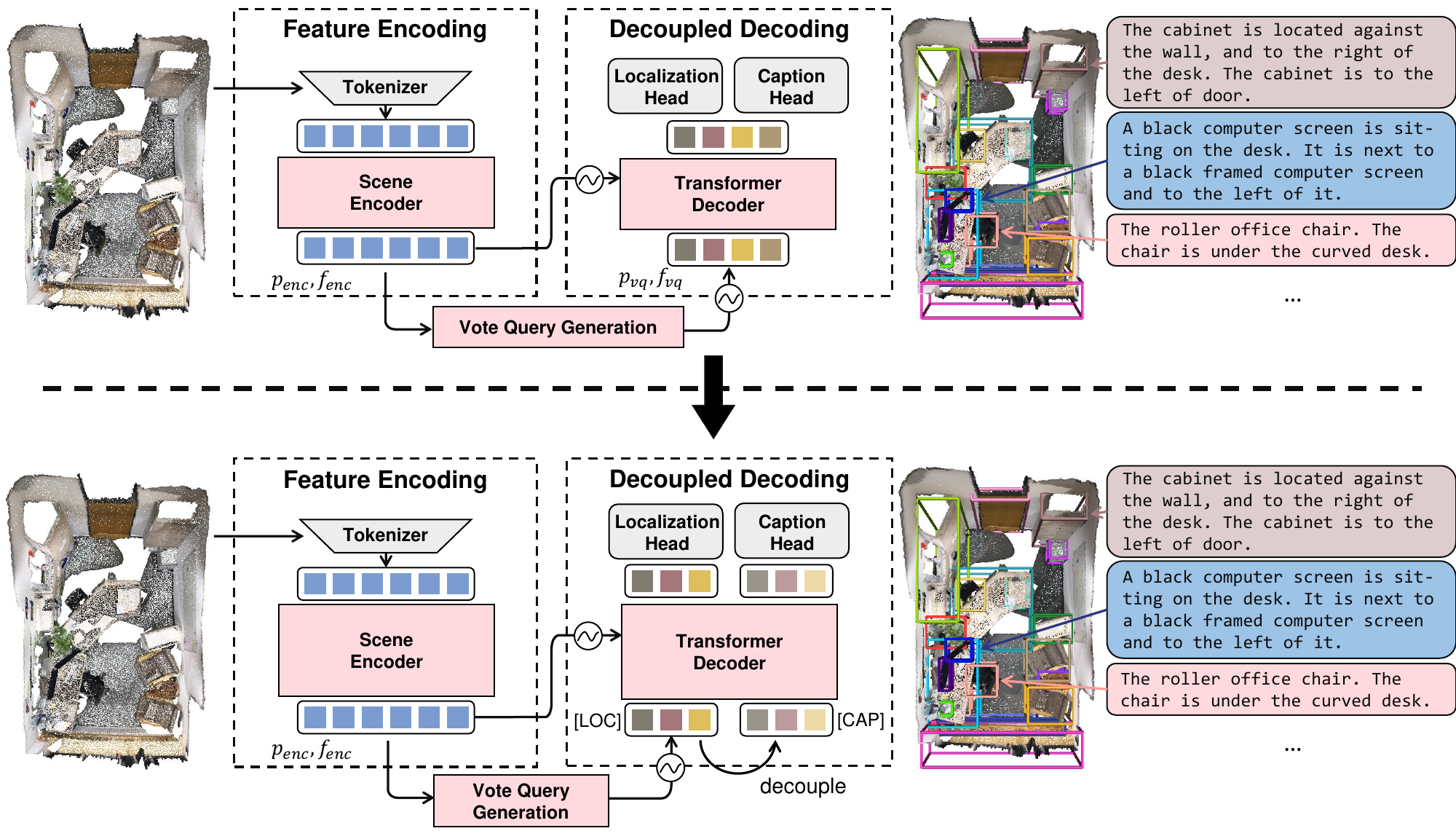}
    \vspace{-5mm}
	\caption{
        \textbf{An overview of our proposed Vote2Cap (upper) and Vote2Cap-DETR++ (bottom) framework.}
        The two proposed frameworks take a 3D point cloud as their input, and generate a set of box-sentence pairs to localize and describe each object in the point cloud.
        The preliminary one (upper) first extracts features of the input 3D scene with a scene encoder.
        Then we generate vote queries $\left(p_{vq}, f_{vq}\right)$ from the encoded scene tokens, and decode the queries into captions and box estimations with two task-specific heads in parallel.
        \whatsnew{
            The advanced model (bottom) differs from the preliminary one mainly in that we decouple the caption queries from the localization queries to capture task-specific features.
            Additionally, we make certain modifications as described in section \ref{subsec:vote2cap-detr++} to the transformer decoder and caption head for tighter box estimations and more informative descriptions.
        }
	}
	\label{fig:pipeline}
\end{figure*}

The \textbf{input} of 3D dense captioning is a 3D scene represented by a set of $N$ points $PC=\left[p_{in}; f_{in}\right]\in \mathbb{R}^{N\times 3+F}$, where $p_{in}\in \mathbb{R}^{N\times 3}$ represents the absolute location, \textit{i.e.} the geometric feature for each point, and $f_{in}\in \mathbb{R}^{N\times F}$ is the additional information for each point, including \textit{color}, \textit{normal}, \textit{height}, and \textit{multiview feature} introduced in \cite{chen2021scan2cap, paszke2016enet}.
The expected \textbf{output} is composed of a set of $K$ box-caption paired estimations $(\hat{B}, \hat{C}) = \{(\hat{b}_i, \hat{c}_i)\vert i=1,\cdots,K\}$, representing the location and descriptions of a total of $K$ distinctive objects in the input 3D scene.

In this section, we first give a brief background introduction of the transformer architecture\cite{vaswani2017attention} in section \ref{subsec:transformer}.
Then, we introduce our preliminary model, Vote2Cap-DETR, in section \ref{subsec:vote2cap-detr}.
After that, we bring out the advanced version, \textit{i.e.}, Vote2Cap-DETR++ in section \ref{subsec:vote2cap-detr++}, which digs deeper into the vote query design and introduces the instruction tuning strategy.
We also introduce the training objectives for both models in section \ref{subsec:supervision-vote2cap-detr} and section \ref{subsec:supervision-vote2cap-detr++}

\subsection{Background: Transformers}
\label{subsec:transformer}
Since its first appearance\cite{vaswani2017attention}, the transformer architecture has been widely adapted to various applications\cite{brown2020gpt3,devlin2018bert,dosovitskiy2020image,li2023blip}.
A transformer consists of stacked encoder/decoder layers, where each encoder/decoder layer is composed of attention layers, a \textbf{F}eed-\textbf{F}orward \textbf{N}etwork (FFN), and residual connections\cite{he2016resnet,ba2016layernorm}.

\myparagraph{Attention Layer.}
The attention operation requires the input of query $x_{q}\in \mathbb{R}^{n \times d}$, key $x_{k}\in \mathbb{R}^{m \times d}$, and value $x_{v}\in \mathbb{R}^{m \times d}$, where $n$, $m$ represent the number of tokens, and $d$ indicates the feature dimension.
The inputs are first projected with separate and learnable \textbf{F}ully \textbf{C}onnected (FC) layers:
\begin{equation}
    x_q = FC\left(x_q\right), \quad 
    x_k = FC\left(x_k\right), \quad 
    x_v = FC\left(x_v\right).
    \label{eq:feature projection}
\end{equation}
Then, the model calculates the weighting matrix:
\begin{equation}
    \alpha = Softmax\left(\frac{x_{q} \cdot x_{k}^{T}}{\sqrt{d}}\right) \in \mathbb{R}^{n\times m}.
\end{equation}
The updated query feature $x_{q}^{\prime} \in \mathbb{R}^{n \times d}$ is obtained by aggregating feature from $x_{v}$ with a weighted sum:
\begin{equation}
    x_{q}^{\prime} = Attn\left(x_{q}, x_{k}, x_{v}\right) = \alpha \cdot x_{v} \in \mathbb{R}^{n \times d}.
\end{equation}
In practice, researchers adopt the multi-head attention\cite{vaswani2017attention}, where they separate the projected input feature in Equation \ref{eq:feature projection}, into several slices before the attention operation, and concatenate the updated query feature after.

\myparagraph{Transformer Encoder Layer.}
A standard transformer encoder layer consists of an attention layer as well as a \textbf{F}eed \textbf{F}oward \textbf{N}etwork (FFN).
Given the $i$-th encoder layer, the query feature $x_{i}$ is updated through:
\begin{equation}
    \begin{aligned}
        x_{i}^{\prime} &= LN\left(x_{i}\right); \\
        x_{i}^{\prime\prime} &= x_{i} + Attn\left(x_{i}^{\prime}, x_{i}^{\prime}, x_{i}^{\prime}\right); \\
        x_{i+1} &= x_{i} + FFN\left(LN\left(x_{i}^{\prime\prime}\right)\right).
    \end{aligned}
\end{equation}

\myparagraph{Transformer Decoder Layer.}
The transformer decoder layer differs from the encoder layer in that, it requires an additional attention layer to aggregate features from another source of information (denoted as $y$):
\begin{equation}
    \begin{aligned}
        x_{i}^{\prime} &= LN\left(x_{i}\right); \\
        x_{i}^{\prime\prime} &= x_{i} + Attn\left(x_{i}^{\prime}, x_{i}^{\prime}, x_{i}^{\prime}\right); \\
        x_{i}^{\prime\prime\prime} &= x_{i} + Attn\left(LN\left(x_{i}^{\prime\prime}\right), y, y\right); \\
        x_{i+1} &= x_{i} + FFN\left(LN\left(x_{i}^{\prime\prime\prime}\right)\right).
    \end{aligned}
\end{equation}
Here, $y$ is usually shared along all decoder layers.

\subsection{Preliminary Version: Vote2Cap-DETR}
\label{subsec:vote2cap-detr}

In this section, we present the pipeline of Vote2Cap-DETR\cite{chen2023vote2cap} as shown in Figure \ref{fig:pipeline}.
The input $PC$ is first tokenized to $2,048$ point tokens with a set-abstraction layer\cite{qi2017pointnet++} following \cite{misra2021-3detr}.
Then, we feed the point tokens to a scene encoder\cite{misra2021-3detr} to extract scene feature $\left[p_{enc}, f_{enc}\right] \in \mathbb{R}^{1,024\times \left(3 + 256\right)}$.
After that, we generate vote queries $\left[p_{vq}, f_{vq}\right]$ from $\left[p_{enc}, f_{enc}\right]$ as initial object queries for the decoder.
Finally, we adopt a transformer decoder\cite{vaswani2017attention} to capture both query-query and query-scene interactions through self-attention and cross-attention, and decode the query feature to box predictions and descriptions in parallel.

\myparagraph{Scene Encoder.} We adopt the same scene encoder as 3DETR\cite{misra2021-3detr} does, which consists of three identical transformer encoder layers with different masking radius of $\left[0.16, 0.64, 1.44\right]$, and a set-abstraction layer\cite{qi2017pointnet++} to downsample the point tokens between the first two encoder layers.
The output of the scene encoder is 1,024 point tokens $\left[p_{enc}, f_{enc}\right] \in \mathbb{R}^{1,024\times \left(3 + 256\right)}$ uniformly distributed in the input 3D scene.

\myparagraph{Vote Query.} For clarification, we formulate the object queries into a form of $\left(p_{query}, f^{0}_{query}\right)$ to represent the spatial position and the initial feature of object queries.
Thus, the object queries in 3DETR\cite{misra2021-3detr} can be represented as $\left(p_{seed}, \mathbf{0}\right)$, where $p_{seed}$ are seed points sampled uniformly from a 3D scene with \textbf{F}arthest \textbf{P}oint \textbf{S}ampling (FPS), and the initial query feature are zero vectors.
However, because of the cluttered 3D scene and sparse object surfaces, $p_{seed}$ could be far away from the scene objects, resulting in slow convergence to capture discriminative object features with further miss detection.
Prior works show that introducing structure bias to initial object queries, such as anchor points\cite{wang2022anchor} and content token selection\cite{zhang2022dino}, is essential for DETRs.
Thus, we propose the vote queries $\left(p_{vq}, f_{vq}\right)$, which introduce both 3D spatial bias and local content aggregation for faster convergence and better performance.

\begin{figure}[htbp]
	\centering
	\includegraphics[width=\linewidth]{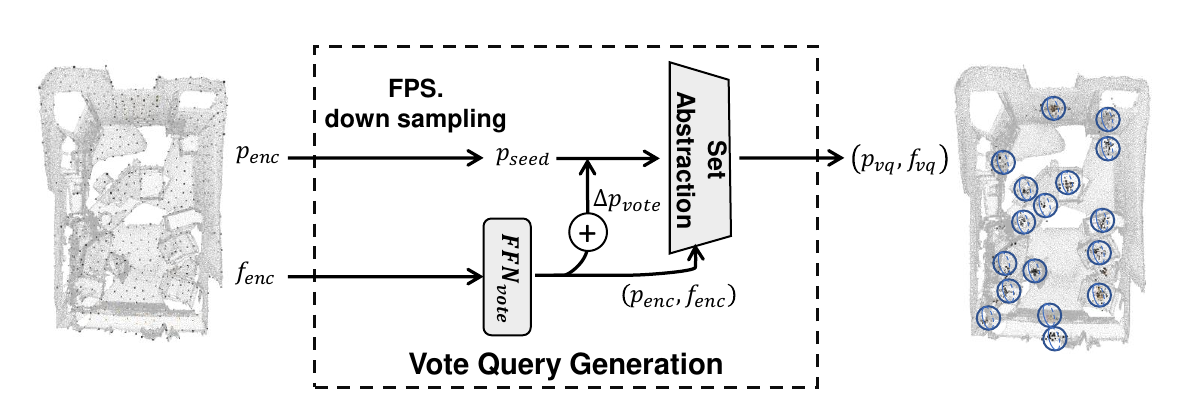}
    \vspace{-5mm}
	\caption{
        \textbf{The detailed generation procedure for vote queries.}
        We first downsample $p_{enc}$ for $p_{seed}$, then we predict the 3D spatial offset $\Delta p_{vote}$ that aims to shift $p_{seed}$ to object centers.
        We achieve $p_{vq} = p_{seed} + \Delta p_{vote}$ and then aggregate local contents for vote query feature $f_{vq}$.
        The vote queries are writtern as $(p_{vq}, f_{vq})$.
	}
	\label{fig:vote query}
\end{figure}

To be specific, the spatial location and initial features of vote queries $p_{vq}, f_{vq}$ are expected to be close to object centers with discriminative representations.
This builds the connection between the object queries and the vote set prediction widely studied in \cite{qi2019votenet}.
The detailed structure can be found in Figure \ref{fig:vote query}.
In practice, we first uniformly sample 256 points from $p_{enc}$ with FPS for seed points $p_{seed}$ as done in 3DETR\cite{misra2021-3detr}.
Then, we predict a 3D spatial offset $\Delta p_{vote}$ from $p_{seed}$'s respective feature $f_{seed}$ with a \textbf{F}eed \textbf{F}orward \textbf{N}etwork (FFN) $FFN_{vote}$ for $p_{vq}$ as shown in Equation \ref{eq:vote_xyz}.
Here, $\Delta p_{vote}$ is trained to estimate the center of nearby objects.
\begin{equation}
    p_{vq} = p_{seed} + \Delta p_{vote} = p_{seed} + FFN_{vote}\left(f_{seed}\right).
    \label{eq:vote_xyz}
\end{equation}
After that, we aggregate the content feature from the scene feature $\left(p_{enc}, f_{enc}\right)$ for $f_{vq}\in \mathbb{R}^{256\times 256}$, with a set abstraction layer\cite{qi2017pointnet++} as done in \cite{qi2019votenet}.
It is also worth mentioning that the seed queries in 3DETR are also a special case in vote queries, where $\Delta p_{vote}=\mathbf{0}$ and $f_{vq}=\mathbf{0}$.

Following 3DETR\cite{misra2021-3detr}, we adopt an eight-layer vanilla transformer decoder and update the $i$-th layer's query feature $f^i_{query}$ through following:
\begin{equation}
    f^i_{query} = Layer_{i-1} \left(f^{i-1}_{query} + PE\left(p_{vq}\right)\right),
    \label{eq:query update}
\end{equation}
where $Layer_{i-1}$ is the $i$-th decoder layer, $PE(\cdot)$ is the 3D Fourier positional encoding function\cite{tancik2020fourier}, and $f^0_{query}=f_{vq}$ as stated above.

\begin{figure}[htbp]
	\centering
	\includegraphics[width=\linewidth]{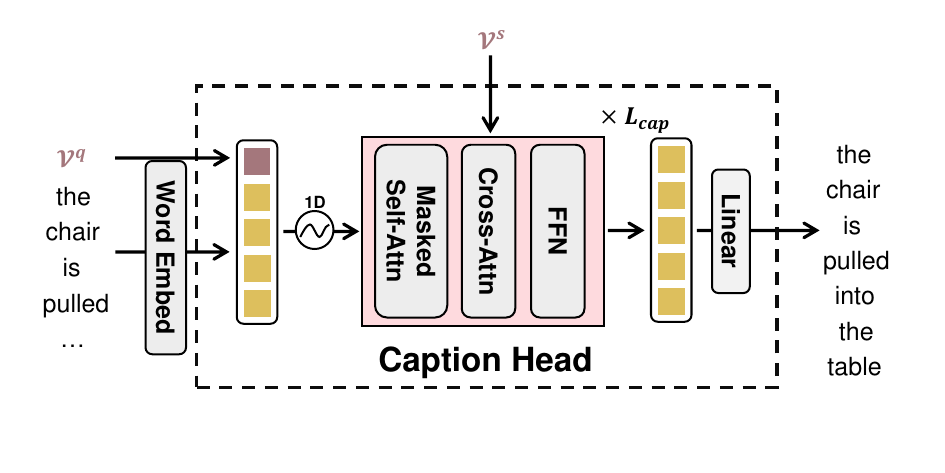}
    \vspace{-1cm}
	\caption{
        \textbf{The details of the Dual Clued Captioner (DCC).}
        To generate informative and object-centric descriptions without visiting the output of the box estimation branch, we utilize two streams of visual clues, \textit{i.e.} the query feature $\mathcal{V}^q$ representing the object to be described, and the query's nearest local contextual tokens $\mathcal{V}^s$.
	}
	\label{fig:captioner}
\end{figure}

\myparagraph{Decoupled and Parallel Decoding.}
We decode object queries concurrently to box estimations and captions with two parallel task-specific heads, the detection head and the caption head. 
It is worth mentioning that these two heads are agnostic to each other's output.
\textbf{\textit{Detection Head}.}
We adopt shared FFNs along all the decoder layers for box corner estimations $\hat{B}$ and semantic category prediction $\hat{S}$ (containing ``no object'' class) following \cite{carion2020detr,misra2021-3detr}.
\textbf{\textit{Caption Head}.}
To generate object-centric and informative descriptions without visiting the box estimations, we propose the \textbf{D}ual-\textbf{C}lued \textbf{C}aptioner (DCC), which is a two-layer transformer decoder\cite{vaswani2017attention} with sinusoid position embedding.
To be specific, DCC receives two streams of visual clue $\mathcal{V} = \left(\mathcal{V}^{q}, \mathcal{V}^{s}\right)$, where $\mathcal{V}^{q}$ represents the query feature of the last decoder layer and $\mathcal{V}^{s}$ is the nearest local context token features surrounding the spatial location of $\mathcal{V}^{q}$.
The caption generation process could be treated as maximizing a conditional probability:
\begin{equation}
    c^{*} = \mathop{\arg\max}_{c} P\left(c \vert \mathcal{V}^{s}; \mathcal{V}^{q}\right),
    \label{eq:caption generation}
\end{equation}
where $c^{*}$ is the caption with the highest conditional probability to be generated.

\subsection{Advanced Version: Vote2Cap-DETR++}
\label{subsec:vote2cap-detr++}

\whatsnew{
To further drive the evolution of ``non-detect-then-describe'' methods for 3D dense captioning, we introduce Vote2Cap-DETR++ (Figure \ref{fig:pipeline}, bottom).
The main difference between the two versions is that we decouple the queries in Vote2Cap-DETR++ to capture task-specific features for the localization head and caption head.
Besides, we apply an iterative spatial refinement strategy on vote queries for better localizing objects in 3D space, and inject additional 3D spatial information for more accurate captions.
}

\begin{figure}[htbp]
	\centering
	\includegraphics[width=\linewidth]{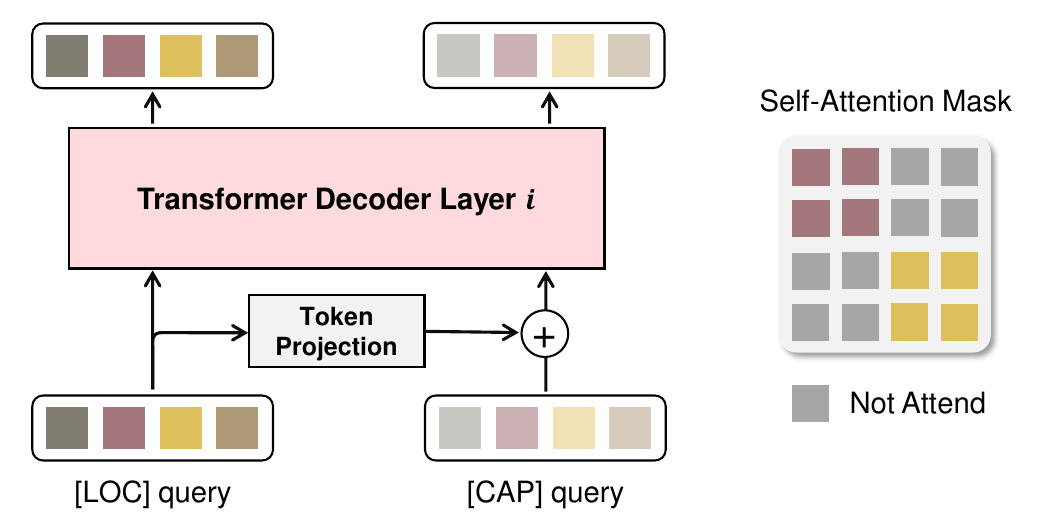}
    \vspace{-5mm}
	\caption{
        \textbf{Decouple-and-Correspond design for task-specific queries.}
        We decouple the queries for object localization (``[LOC]'') and caption generation (``[CAP]'') to capture task-specific features, and link the queries from the same spatial position via token-wise projection.
	}
	\label{fig:query-design}
\end{figure}

\myparagraph{Decoupling Task-Specific Queries.} 
\whatsnew{
Previous ``detect-then-describe'' methods\cite{chen2021scan2cap,wang2022spacap3d} adopt it as a \textit{de-facto} standard to generate captions with object proposal features.
The above-introduced Vote2Cap-DETR also shares the same set of vote queries for simultaneous object localization and caption generation for scene objects.
However, these two sub-tasks necessitate different levels of understanding of the 3D environment. 
The former requires a model to perceive 3D structures for tight box estimations, whereas the latter calls for sufficient attribute information along with spatial relations to the surroundings.
}

\whatsnew{
Thus, we propose the decoupled-and-correspond queries to capture task-specific features via the transformer decoder as is shown in Figure \ref{fig:pipeline} (bottom).
For clarification, we name the first set of queries as ``[LOC]'' queries for object localization and the second set as ``[CAP]'' queries to capture features for caption generation. 
\textbf{Decoupling.} The ``[LOC]'' queries are indeed the vote queries propose in section \ref{subsec:vote2cap-detr}. 
To distinguish the two queries, we project the feature of a ``[LOC]'' query for the corresponding ``[CAP]'' query while their spatial location is shared.
\textbf{Correspondence.} Though the two sets of task-specific queries are designed to have a different understanding of a 3D scene, each paired queries share the same spatial location and are tied to the same box-caption proposal.
Therefore, we link the tokens in [LOC] query and [CAP] query in each decoder layer via token-wise projection as is shown Figure \ref{fig:query-design}.
}

\begin{figure}[htbp]
    \centering
    \begin{minipage}[t]{0.48\linewidth}
        \centering
        \includegraphics[width=\linewidth]{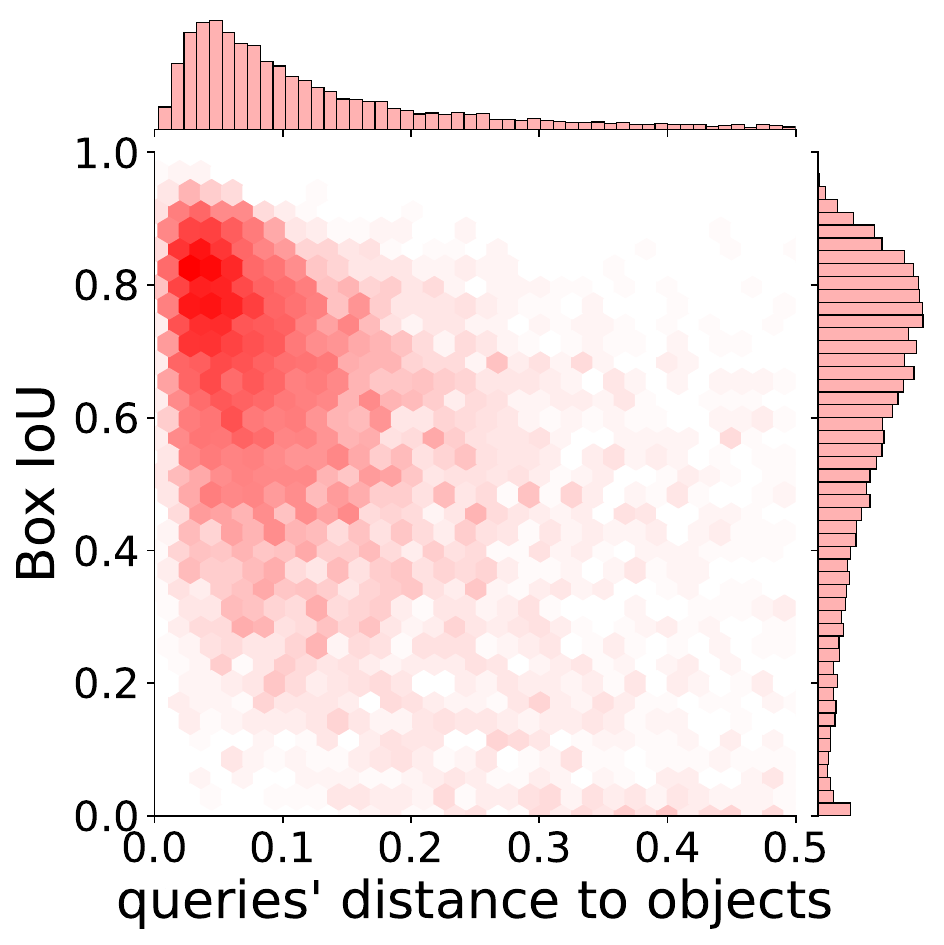}
    \end{minipage}
    \hfill
    \begin{minipage}[t]{0.48\linewidth}
        \centering
        \includegraphics[width=\linewidth]{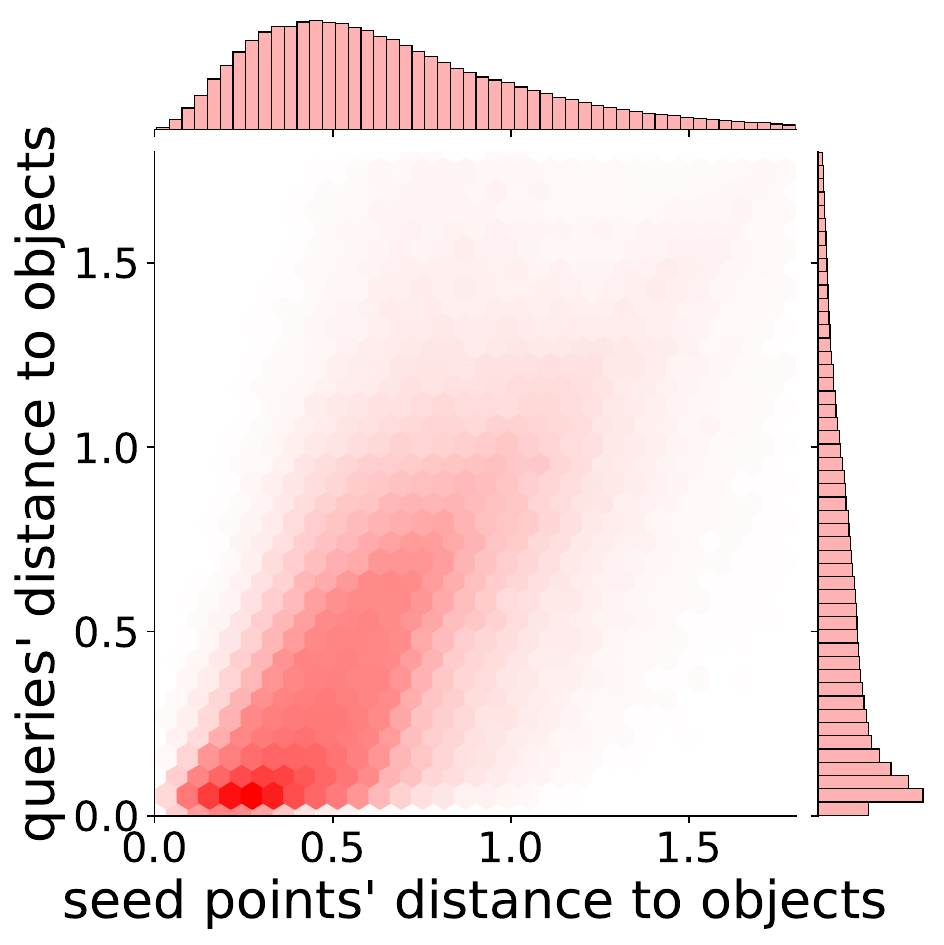}
    \end{minipage}
    \caption{
        \textbf{The importance of vote queries' spatial location.}
        The scatter plot represents the joint distribution of the two variables, while the histograms represent the distribution of each variable.
        The closer a query is to an object, the higher probability it can be decoded to a tight bounding box (left).
        It is hard to shift a seed point to object centers especially when it is far from the objects (right).
	}
    \label{fig:query quality}
\end{figure}

\myparagraph{Iterative Spatial Refinement on Vote Query.}
\whatsnew{
To further unleash the power of vote queries, we visualize the relationship between a query's spatial location and the quality of its box estimation in Figure \ref{fig:query quality} (left).
One can see that the closer a query is to an object, the higher quality a box estimation has.
However, learning to shift seed points to object centers might be challenging, especially for the queries that are initially far from the objects (Figure \ref{fig:query quality}, right).
}

\whatsnew{
To alleviate this issue, we propose to update the spatial location of the vote queries along with the feature updating in Equation \ref{eq:query update}.
Specifically, for the $i$-th decoder layer, we predict a spatial refinement offset $\Delta p_{vote}^{i}$ via an additional FFN.
The query feature updating step in the $i$-th layer could be written as:
\begin{equation}
    \begin{aligned}
        f^i_{query} &= Layer_{i-1} \left(f^{i-1}_{query} + PE\left(p^i_{vq}\right)\right);\\
        p^i_{vq} =& p^{i-1}_{vq} + \Delta p_{vote}^{i - 1} = p^{i-1}_{vq} + FFN\left(f^i_{query}\right)
    \end{aligned}
\end{equation}
It is worth mentioning that we still adopt the decoupled decoding structure to bypass the cumulative error brought up by the ``detect-then-describe'' pipeline, which differs from the iterative box refinement strategy proposed in \cite{zhang2022dino,zhu2020deformabledetr}.
}
\suspicious{
Our method also differs from \cite{zhao2021transformer3d} in that we mainly focus on the improvement of object queries in the DETR-like architecture, while \cite{zhao2021transformer3d} builds the vote refinement on VoteNet\cite{qi2019votenet} for better proposals.
}

\myparagraph{Injecting Spatial Information to the Caption Head.}
\whatsnew{
An informative scene object description may contain terms like ``corner of the room'', ``middle of the room'', etc.
However, the original DCC in Figure \ref{fig:captioner} is not capable of capturing sufficient absolute spatial information.
Thus, we insert an additional 3D absolute position token $\mathcal{V}^{q}_{pos}$ to the caption prefix to identify a query's spatial location.
The caption prefix could be written as $\left[\mathcal{V}^{q}; \mathcal{V}^{q}_{pos}\right]$.
It is worth mentioning that in Vote2Cap-DETR++, $\mathcal{V}^{q}$ comes from the ``[CAP]'' queries rather than the ``[LOC]'' queries in the preliminary model.
To further inform the caption head of the relation between a local context token and a caption query, we adopt a ranking-based position embedding for the context tokens with respect to their spatial distance to the query.
We encode the context tokens' spatial position as $\mathcal{V}^{s}_{pos}$ with the same sinusoid position embedding added to the word embedding for caption generation.
Therefore, the conditional description generation process in Equation \ref{eq:caption generation} can be reformulated as:
\begin{equation}
    c^{*} = \mathop{\arg\max}_{c} P\left(c \vert \mathcal{V}^{s}; \mathcal{V}^{s}_{pos}; \mathcal{V}^{q}; \mathcal{V}^{q}_{pos}\right).
    \label{eq:caption generation++}
\end{equation}
}

\subsection{Training Objective for Vote2Cap-DETR}
\label{subsec:supervision-vote2cap-detr}
The loss function of Vote2Cap-DETR is a weighted sum of a total of three losses: the vote query loss $\mathcal{L}_{vq}$, the detection loss $\mathcal{L}_{det}$, and the caption loss $\mathcal{L}_{cap}$.

\myparagraph{Vote Query Loss.}
In practice, we supervise vote shifting procedure with all 1,024 points in $p_{enc}$ for $p_{vote}$ with the same procedure in Equation \ref{eq:vote_xyz}, where $p_{seed}$ is 256 points sampled from $p_{enc}$.
We adopt the vote loss proposed in VoteNet\cite{qi2019votenet} as $\mathcal{L}_{vq}$ to facilitate the learning of shifting points towards object centers: 
\begin{equation}
    \mathcal{L}_{vq} = \frac{1}{M}
        \sum_{i = 1}^{M} \sum_{j = 1}^{N_{gt}}
            \left\|p_{vote}^{i} - cnt_{j}\right\|_{1}
                \cdot 
                \mathbb{I}\left\{p_{enc}^{i} \in I_j\right\}.
    \label{eq:loss vote query}
\end{equation}
Herein, $\mathbb{I}(\cdot)$ is an indicator function that takes the value of $1$ when the condition meets and $0$ otherwise. 
The variable $N_{gt}$ represents the number of instances present in a 3D scene, $M$ calculates the number of points in $p_{vote}$, which is equal to 1,024 in our setting.
Finally, $cnt_{j}$ denotes the center of the $j$th instance, denoted as $I_j$.

\myparagraph{Detection Loss.}
We adopt the same Hungarian algorithm as 3DETR\cite{misra2021-3detr} to assign each proposal with a ground truth label. 
We involve a larger weight on the 3D gIoU loss\cite{misra2021-3detr} to enhance the model's object localization capability:

\begin{equation}
    \mathcal{L}_{set} = 
        \alpha_1 \mathcal{L}_{giou}
        + \alpha_2 \mathcal{L}_{cls}
        + \alpha_3 \mathcal{L}_{center-reg} 
        + \alpha_4 \mathcal{L}_{size-reg},
\end{equation}
where $\alpha_1=10$, $\alpha_2=1$, $\alpha_3=5$, $\alpha_4=1$ are set heuristically.
Moreover, the set loss $\mathcal{L}_{set}$ is along all $n_{dec-layer}$ decoder layers\cite{misra2021-3detr}.

\myparagraph{Caption Loss.}
Following the standard practice of image captioning, we first train our caption head with the standard cross-entropy loss (MLE training) and then fine-tune it with \textbf{S}elf-\textbf{C}ritical \textbf{S}equence \textbf{T}raining (SCST)\cite{rennie2017scst}.
During MLE training, the model is trained to predict the $\left(t+1\right)$-th word $c_i^{t+1}$, given the first $t$ words $c_i^{[1:t]}$ and the visual condition $\mathcal{V}$.
The loss function for a $T$-length sentence can be defined as:
\begin{equation}
    \mathcal{L}_{c_i} = \sum_{i=1}^{T} \mathcal{L}_{c_i}(t) = -\sum_{i=1}^{T} \log \hat{P}\left(c_i^{t+1} \vert \mathcal{V}; c_i^{[1:t]}\right).
    \label{eq:cap-mle}
\end{equation}
After training the caption head under word-level supervision, we fine-tune it with SCST. 
During SCST, the model generates multiple captions $\hat{c}_{1, \cdots,k}$ using beam search with a beam size of $k$, and also generates a baseline caption $\hat{g}$ using greedy search. 
The loss function for SCST is defined as follows:
\begin{equation}
    \mathcal{L}_{c_i} = 
    - \sum_{i=1}^{k}
        \left(R\left(\hat{c}_{i}\right) - R\left(\hat{g}\right)\right) 
        \cdot
        \frac{1}{\left|\hat{c}_i\right|}\log \hat{P}\left(\hat{c}_i\vert \mathcal{V}\right)
        .
    \label{eq:cap-scst}
\end{equation}
Here, the reward function $R\left(\cdot\right)$ in our case is CIDEr\cite{vedantam2015cider}, which is commonly used to evaluate the text generation models.
To encourage equal importance among captions of different lengths, we normalize the log probability of caption $\hat{c}_i$ by its length, which is denoted as $\left|\hat{c}_i\right|$.

\myparagraph{Set to Set Training for 3D Dense Captioning.}
We introduce an easy-to-implement set-to-set training strategy for 3D dense captioning.
Specifically, given a 3D scene, we randomly sample one sentence from the corpus per annotated instance.
Following that, we assign each instance's language annotation to one distinctive proposal in the corresponding scene with the same Hungarian algorithm.
During training, we average the losses for captions $\mathcal{L}_{c_i}$ on all annotated instances within a batch to compute the caption loss $\mathcal{L}_{cap}$.
To balance losses for different tasks, we adopt a weighted sum along all loss functions during training:
\begin{equation}
    \mathcal{L}_\text{\scriptsize{Vote2Cap-DETR}} = \beta_1\mathcal{L}_{vq} + \beta_2 \sum_{i=1}^{n_{dec-layer}}\mathcal{L}_{set} + \beta_3 \mathcal{L}_{cap},
    \label{eq:loss_vote2cap_DETR}
\end{equation}
where $\beta_1 = 10$, $\beta_2 = 1$, $\beta_3 = 5$ are set heuristically.

\subsection{Training Objective in Vote2Cap-DETR++}
\label{subsec:supervision-vote2cap-detr++}

\myparagraph{Spatial Refinement Loss for Queries.}
\whatsnew{
In Vote2Cap-DETR++, we further adopt a refinement loss $\mathcal{L}_{qr}$ for queries in different decoder layers.
$\mathcal{L}_{qr}$ has a similar form with $\mathcal{L}_{vq}$ defined in Equation \ref{eq:loss vote query}, but is only adopted to the vote queries:
\begin{equation}
    \mathcal{L}_{qr} = \frac{1}{M}
    \sum_{i = 1}^{M} \sum_{j = 1}^{N_{gt}}
        \left\|p_{vq}^{i} - cnt_{j}\right\|_{1}
            \cdot 
            \mathbb{I}\left\{p_{vq}^{i} \in I_j\right\}.
\end{equation}
Here, $p_{vq}^{i}$ is the spatial location of queries in the $i$-th decoder layer, while other notations are defined in Equation \ref{eq:loss vote query} accordingly.
We adopt $\mathcal{L}_{qr}$ for each decoder layer that refines the spatial location of vote queries.
}

\myparagraph{Loss Function for Vote2Cap-DETR++.}
\whatsnew{
The final loss function for Vote2Cap-DETR++ builds upon $\mathcal{L}_\text{\scriptsize{Vote2Cap-DETR}}$ introduced in Equation \ref{eq:loss_vote2cap_DETR}, but further takes account into the above mentioned refinement loss $\mathcal{L}_{qr}$:
\begin{equation}
    \mathcal{L}_\text{\scriptsize{Vote2Cap-DETR++}} = \mathcal{L}_\text{\scriptsize{Vote2Cap-DETR}} + \beta_4 \sum_{i \in \delta}\mathcal{L}_{qr},
\end{equation}
where $\delta$ stands for all the decoder layers that perform spatial refinement for queries.
We empirically set $\beta_4=\beta_1 = 10$.
}
\section{Experiments}
\label{sec:exp}

In this section, we first introduce basic settings in 3D dense captioning, including the datasets, metrics, and implementation details in section \ref{subsec:datasets,metric,implementation}.
Then, we compare the two proposed methods with previous state-of-the-art approaches in section \ref{subsec:comparison with existing}.
After that, we provide ablation studies on Vote2Cap-DETR and Vote2Cap-DETR++ 
 in section \ref{subsec:ablations-vote2cap} and section \ref{subsec:ablations-vote2cap++}.
Finally, we provide some qualitative results in section \ref{subsec:viz}.

\subsection{Datasets, Metrics, and Implementation Details}
\label{subsec:datasets,metric,implementation}

\myparagraph{Datasets}.
We conduct experiments on two widely used datasets for 3D dense captioning, namely ScanRefer\cite{chen2020scanrefer} and Nr3D\cite{achlioptas2020referit3d}.
ScanRefer/Nr3D contains 36,665/32,919 human-annotated natural language descriptions on 7,875/4,664 objects from 562/511 out of 1201 3D scenes in ScanNet\cite{dai2017scannet} for training, and 9,508/8,584 descriptions on 2,068/1,214 objects from 141/130 out of 312 3D scenes from the ScanNet validation set for evaluation.

\myparagraph{Evaluation Metrics}.
Though our proposed method is robust to NMS\cite{neubeck2006nms}, we follow the same procedure in \cite{chen2021scan2cap,chen2023vote2cap} to obtain the final predictions by applying NMS on the model's box-caption predictions for a fair comparison.
After that, we assign each instance annotation with an object-caption proposal from the remaining set with the largest IoU.
Here, we use $(b_i, C_i)$ to represent the annotation for each instance, where $b_i$ is an instance's box corner label, and $C_i$ is the corpus containing all caption annotations for this instance.
To jointly evaluate the model's localization and caption generation capability, we adopt the $m@kIoU$ metric\cite{chen2021scan2cap}:
\begin{equation}
    m@kIoU=\frac{1}{N}\sum_{i=1}^{N} m\left(\hat{c}_i, C_i\right) \cdot \mathbb{I}\left\{IoU\left(\hat{b}_i, b_i\right) \ge k\right\}.
\label{eq:m@kIoU}
\end{equation}
Here, $N$ is the number of all annotated instances in the evaluation dataset, and $m$ could be any metric among CIDEr\cite{vedantam2015cider}, METEOR\cite{banerjee2005meteor}, BLEU-4\cite{papineni2002bleu}, and ROUGE-L\cite{lin2004rouge}.

\begin{table*}[htbp]
    \caption{
        \textbf{Quantitative comparisons on the ScanRefer\cite{chen2020scanrefer} validation set.}
        We follow the exact and standard evaluation protocol from Scan2Cap\cite{chen2021scan2cap}, and make separate comparisons according to different caption supervisions (MLE Training and SCST introduced in section \ref{subsec:supervision-vote2cap-detr}) with all published state-of-the-art 3D dense captioning methods.
        The methods marked $^{*}$ are trained with extra data.
        Our proposed methods achieve a new state-of-the-art.
    }
    \label{tab:scanrefer}
    \centering
    \resizebox{\linewidth}{!}{
    \begin{tabular}{cccccccccccccccccccccc}
    \toprule
    \multirow{3}{*}{Method} & \multirow{3}{*}{$\mathcal{L}_{des}$} &  & \multicolumn{9}{c}{w/o additional 2D input}                                                                          &  & \multicolumn{9}{c}{w/ additional 2D input}                                                                            \\
                            &                                      &  & \multicolumn{4}{c}{IoU = 0.25}                          &  & \multicolumn{4}{c}{IoU = 0.50}                          &  & \multicolumn{4}{c}{IoU = 0.25}                          &  & \multicolumn{4}{c}{IoU = 0.50}                          \\ \cline{4-7} \cline{9-12} \cline{14-17} \cline{19-22} 
                            &                                      &  & C$\uparrow$ & B-4$\uparrow$ & M$\uparrow$ & R$\uparrow$ &  & C$\uparrow$ & B-4$\uparrow$ & M$\uparrow$ & R$\uparrow$ &  & C$\uparrow$ & B-4$\uparrow$ & M$\uparrow$ & R$\uparrow$ &  & C$\uparrow$ & B-4$\uparrow$ & M$\uparrow$ & R$\uparrow$ \\ \hline
    Scan2Cap\cite{chen2021scan2cap}                & \multirow{12}{*}{MLE}                 &  & 53.73       & 34.25         & 26.14       & 54.95       &  & 35.20       & 22.36         & 21.44       & 43.57       &  & 56.82       & 34.18         & 26.29       & 55.27       &  & 39.08       & 23.32         & 21.97       & 44.78       \\
    MORE\cite{jiao2022more}                    &                                      &  & 58.89       & 35.41         & 26.36       & 55.41       &  & 38.98       & 23.01         & 21.65       & 44.33       &  & 62.91       & 36.25         & 26.75       & 56.33       &  & 40.94       & 22.93         & 21.66       & 44.42       \\
    SpaCap3d\cite{wang2022spacap3d}   &                                      &  & -           & -             & -           & -           &  & 42.76       & 25.38         & 22.84       & 45.66       &  & -           & -             & -           & -           &  & 44.02       & 25.26         & 22.33       & 45.36       \\
    REMAN\cite{mao2023complete}       &                                      &  & -           & -             & -           & -           &  & -           & -             & -           & -           &  & 62.01       & 36.37         & 26.76       & 56.25       &  & 45.00       & 26.31         & 22.67       & 46.96       \\
    D3Net\cite{chen2021d3net}         &                                      &  & -           & -             & -           & -           &  & -           & -             & -           & -           &  & -           & -             & -           & -           &  & 46.07       & 30.29         & 24.35       & 51.67       \\
    Contextual\cite{zhong2022contextual3DdenseCap} &                                      &  & -       & -        & -       & -          &  & 42.77       & 23.60         & 22.05       & 45.13       &  & -       & -         & -       & -       &  & 46.11       & 25.47         & 22.64       & 45.96       \\
    UniT3D$^{*}$\cite{chen2022unit3d}         &                                      &  & -           & -             & -           & -           &  & -           & -             & -           & -           &  & -           & -             & -           & -           &  & 46.69       & 27.22         & 21.91       & 45.98       \\
    3DJCG\cite{cai20223djcg}                   &                                      &  & 60.86       & 39.67         & 27.45       & 59.02       &  & 47.68       & 31.53         & 24.28       & 51.80       &  & 64.70       & 40.17     & 27.66       & 59.23       &  & 49.48       & 31.03         & 24.22       & 50.80       \\
    %
    %
    3D-VLP$^{*}$\cite{jin2023context}      &                                      &  & 64.09        & 39.84        & 27.65       & 58.78          &  & 50.02       & 31.87         & 24.53       & 51.17       &  & 70.73       & \textbf{41.03}         & 28.14       & \textbf{59.72}       &  & 54.94       & 32.31         & 24.83       & 51.51       \\
    \cline{1-1}
    \multicolumn{1}{l}{\textbf{\textit{Ours:}}} \\
    Vote2Cap-DETR\cite{chen2023vote2cap} &                                      
        &  & 71.45   & 39.34   & 28.25   & 59.33
        &  & 61.81   & 34.46   & 26.22   & 54.40
        &  & 72.79   & 39.17   & 28.06   & 59.23
        &  & 59.32   & 32.42   & 25.28   & 52.53       \\ 
    Vote2Cap-DETR++                      &                                      
        &  & \textbf{76.36}   & \textbf{41.37}   & \textbf{28.70}   & \textbf{60.00}
        &  & \textbf{67.58}   & \textbf{37.05}   & \textbf{26.89}   & \textbf{55.64}
        &  & \textbf{77.03}   & 40.99   & \textbf{28.53}   & 59.59   
        &  & \textbf{64.32}   & \textbf{34.73}   & \textbf{26.04}   & \textbf{53.67}  \\
    \hline
    $\chi$-Trans2Cap\cite{yuan2022x-trans2cap}        & \multirow{7}{*}{SCST}                &  & 58.81       & 34.17         & 25.81       & 54.10       &  & 41.52       & 23.83         & 21.90       & 44.97       &  & 61.83       & 35.65         & 26.61       & 54.70       &  & 43.87       & 25.05         & 22.46       & 45.28       \\
    Scan2Cap\cite{chen2021scan2cap}                &                                      &  & -           & -             & -           & -           &  & -             & -             & -           & -           &  & -           & -             & -           & -           &  & 48.38       & 26.09         & 22.15       & 44.74       \\
    Contextual\cite{zhong2022contextual3DdenseCap} &                                      &  & -       & -        & -       & -          &  & 50.29       & 25.64         & 22.57       & 44.71       &  & -       & -         & -       & -       &  & 54.30       & 27.24         & 23.30       & 45.81       \\
    D3Net\cite{chen2021d3net}                   &                                      &  & -           & -             & -           & -           &  & -           & -           & -           & -           &  & -           & -             & -           & -           &  & 62.64       & 35.68         & 25.72       & 53.90       \\
    \cline{1-1}
    \multicolumn{1}{l}{\textbf{\textit{Ours:}}} \\
    Vote2Cap-DETR\cite{chen2023vote2cap}           &                                &  & 84.15   & 42.51   & 28.47   & \textbf{59.26}     
        &  & 73.77   & 38.21   & 26.64   & 54.71
        &  & 86.28   & 42.64   & 28.27   & 59.07   
        &  & 70.63   & 35.69   & 25.51   & 52.28    \\ 
    Vote2Cap-DETR++                      &                                      
        &  & \textbf{88.28}   & \textbf{44.07}   & \textbf{28.75}   & \textbf{59.89}
        &  & \textbf{78.16}   & \textbf{39.72}   & \textbf{26.94}   & \textbf{55.52}
        &  & \textbf{88.56}   & \textbf{43.30}   & \textbf{28.64}   & \textbf{59.19}   
        &  & \textbf{74.44}   & \textbf{37.18}   & \textbf{26.20}   & \textbf{53.30}  \\
    \bottomrule
    \end{tabular}
    }
    \label{exp:comparison_on_scanrefer}
\end{table*}
\begin{table}[htbp]
    \centering
    \caption{
        \textbf{Quantitative comparisons on the Nr3D\cite{achlioptas2020referit3d} validation set.}
        Likewise, we train and evaluate our methods on the Nr3D dataset.
        The proposed methods surpass prior arts under both MLE training and SCST.
    }
    \label{tab:nr3d}
    \resizebox{\linewidth}{!}{
    \begin{tabular}{cccccc}
    \toprule
    Method          & $\mathcal{L}_{des}$   & C@0.5$\uparrow$ & B-4@0.5$\uparrow$ & M@0.5$\uparrow$ & R@0.5$\uparrow$ \\ \hline
    Scan2Cap\cite{chen2021scan2cap}        & \multirow{9}{*}{MLE}  & 27.47           & 17.24             & 21.80           & 49.06           \\
    SpaCap3d\cite{wang2022spacap3d}        &                       & 33.71           & 19.92             & 22.61           & 50.50           \\
    D3Net\cite{chen2021d3net}           &                       & 33.85           & 20.70             & 23.13           & 53.38           \\
    REMAN\cite{mao2023complete}         &                       & 34.81           & 20.37             & 23.01           & 50.99           \\
    Contextual\cite{zhong2022contextual3DdenseCap}&                       & 35.26           & 20.42             & 22.77           & 50.78           \\
    3DJCG\cite{cai20223djcg}           &                       & 38.06           & 22.82             & 23.77           & 52.99           \\
    \cline{1-1}
    \multicolumn{1}{l}{\textbf{\textit{Ours:}}} \\
    Vote2Cap-DETR            &                       & 43.84  & 26.68    & 25.41  & 54.43  \\ 
    Vote2Cap-DETR++          &                       & \textbf{47.08}   & \textbf{27.70}   & \textbf{25.44}   & \textbf{55.22} \\
    \hline
    %
    %
    %
    $\chi$-Tran2Cap\cite{yuan2022x-trans2cap} & \multirow{6}{*}{SCST} & 33.62           & 19.29             & 22.27           & 50.00           \\
    Contextual\cite{zhong2022contextual3DdenseCap}&                       & 37.37           & 20.96             & 22.89           & 51.11           \\
    D3Net\cite{chen2021d3net}           &                       & 38.42           & 22.22             & 24.74           & 54.37           \\
    \cline{1-1}
    \multicolumn{1}{l}{\textbf{\textit{Ours:}}} \\
    Vote2Cap-DETR            &                       & 45.53  & 26.88    & 25.43  & 54.76  \\ 
    Vote2Cap-DETR++          &                       & \textbf{47.62}   & \textbf{28.41}   & \textbf{25.63}   & \textbf{54.77} \\
    \bottomrule
    \end{tabular}
    }
\end{table}

\myparagraph{Implementation Details}.
We provide details for different baseline implementations. 
``w/o additional 2D'' refers to the case that the input point cloud $\mathcal{PC}\in \mathbb{R}^{40,000 \times 10}$ contains the absolute spatial location as well as \textit{color}, \textit{normal} and \textit{height} for $40,000$ points representing a 3D scene.
``additional 2D'' replaces the \textit{color} information in the above case with a $128$-dimensional \textit{multiview} feature extracted by ENet\cite{chen2020hgnet} from multi-view images following \cite{chen2021scan2cap}.

We first pre-train the whole network without the caption head on the ScanNet\cite{dai2017scannet} training set for $1,080$ epochs, which is about 163k iterations ($\sim$34 hours).
To train the model, we use an AdamW optimizer\cite{loshchilov2017AdamW} with a learning rate decaying from $5\times 10^{-4}$ to $10^{-6}$ by a cosine annealing scheduler, a weight decay of $0.1$, a gradient clipping of $0.1$, and a batch size of $8$ following \cite{misra2021-3detr}.
Then, we load the pre-trained weights and jointly train the full model with the MLE caption loss in Equation \ref{eq:cap-mle} for another 720 epochs, which is about 51k and 46k iterations for ScanRefer ($\sim$11 hours) and Nr3D ($\sim$10 hours) respectively.
To prevent overfitting, we fix the learning rate of all parameters in the backbone as $10^{-6}$ and set that of the caption head decaying from $10^{-4}$ to $10^{-6}$ with a cosine annealing scheduler.
During SCST, we tune the caption head with a batch size of 2 for 180 epochs with a frozen backbone because of a high memory cost.
This training procedure takes 50k and 46k iterations for ScanRefer ($\sim$14 hours) and Nr3D respectively ($\sim$11 hours) with a fixed learning rate of $10^{-6}$.
We evaluate the model every $2,000$ iterations during training for consistency with existing works\cite{chen2021scan2cap,wang2022spacap3d}, and all experiments mentioned above are conducted on a single RTX3090 GPU.

\subsection{Comparison with Existing Methods}
\label{subsec:comparison with existing}
We compare both of our proposed methods, Vote2Cap-DETR and Vote2Cap-DETR++, with prior arts on two widely used datasets, namely ScanRefer\cite{chen2020scanrefer} and Nr3D\cite{achlioptas2020referit3d}.
We use \textbf{C}, \textbf{B-4}, \textbf{M}, \textbf{R} as abbreviations for CIDEr\cite{vedantam2015cider}, BLEU-4\cite{papineni2002bleu}, METEOR\cite{banerjee2005meteor}, and Rouge-L\cite{lin2004rouge}, respectively.
We mainly compare the C@0.5 metric on both ScanRefer (Table \ref{tab:scanrefer}) and Nr3D (Table \ref{tab:nr3d}) and sort the results in both tables accordingly. 
In Table \ref{tab:scanrefer}, ``-'' indicates that neither the paper nor any other follow-up works have provided such results.
Since different supervisions have a dramatic influence on the captioning performance, we make separate comparisons for MLE training and \textbf{S}elf-\textbf{C}ritical \textbf{S}equence \textbf{T}raining (SCST).
Among all the listed methods, D3Net\cite{chen2021d3net} and Unit3D\cite{chen2022unit3d} adopt an instance segmentation model, PointGroup\cite{jiang2020pointgroup}, for object localization other than conventional 3D detectors.
3DJCG\cite{cai20223djcg} improves VoteNet's localization performance with an FCOS\cite{tian2019fcos} head, which generates box estimations by predicting spatial distance from a voting point to each side of a 3D bounding box.
Other works all adopt the vanilla VoteNet\cite{qi2019votenet} as their object localization backbone.
Additionally, since prior works including 3DJCG\cite{cai20223djcg}, D3Net\cite{chen2021d3net}, Unit3D\cite{chen2022unit3d} and 3D-VLP\cite{jin2023context} shift their attention to the mutual promotion of different 3DVL tasks and train their models on various tasks, we report their fine-tuned performance on 3D dense captioning in both tables.

\whatsnew{
The evaluations on the ScanRefer validation set (Table \ref{tab:scanrefer}) show that Vote2Cap-DETR and Vote2Cap-DETR++ surpass prior arts.
For example, under MLE training with additional 2D inputs, Vote2Cap-DETR achieves 59.32\% C@0.5 while 3D-VLP\cite{jin2023context} achieves 54.94\% with additional training data. 
Additionally, under SCST, our Vote2Cap-DETR achieves 70.63\% C@0.5, which is \textcolor{mygreen}{+7.99}\% higher than the current state-of-the-art model D3Net\cite{chen2021d3net} (62.64\% C@0.5).
Our advanced model, Vote2Cap-DETR++ further achieves 64.32\% C@0.5 (\textcolor{mygreen}{+5.00}\%) under MLE training and 74.44\% C@0.5 (\textcolor{mygreen}{+3.81}\%) under SCST.
}

\whatsnew{
We also present the evaluation results on the Nr3D validation set in Table \ref{tab:nr3d}.
The reported results for Scan2Cap\cite{chen2021scan2cap} comes from the best-reported
results from \cite{cai20223djcg}. 
Training the model MLE, our proposed Vote2Cap-DETR achieve 43.84\% C@0.5, which is \textcolor{mygreen}{+5.78}\% higher than the current art, 3DJCG (38.06\% C@0.5).
The advanced Vote2Cap-DETR++ further achieves an improvement of \textcolor{mygreen}{+3.24}\% and reaches 47.08\% C@0.5 under the exact same setting.
Under SCST, Vote2Cap-DETR also surpasses the current art (D3Net, 38.42\% C@0.5) by \textcolor{mygreen}{+7.11}\% and reaches 43.84\% C@0.5, while the advanced Vote2Cap-DETR++ further achieves another absolute improvement of \textcolor{mygreen}{+2.09}\% C@0.5 (47.62\% C@0.5).
}

\subsection{Ablation Studies on Vote2Cap-DETR}
\label{subsec:ablations-vote2cap}

We conduct extensive experiments with Vote2Cap-DETR to study the effectiveness of the proposed components.
Without further specification, all experiments are conducted under the ``w/o additional 2D'' setting introduced in section \ref{subsec:datasets,metric,implementation}.

\begin{table}[htb]
    \centering
    \caption{
        \textbf{Localization performance of different query designs in the $20k$-th, $40k$-th, $80k$-th, and $160k$-th iteration on the ScanNet validation set.} 
        Introducing $p_{vq}$ as query positions leads to a 0.97\% mAP@0.5$\uparrow$ than the 3DETR-m\cite{misra2021-3detr} baseline.
        Besides, aggregating query feature $f_{vq}$ from its local content for initial query features leads to a boost in both performance and convergence.
    }
    \label{tab:ablation_detection}
    \resizebox{\linewidth}{!}{
    \begin{tabular}{ccccccccc}
    \toprule
    \multicolumn{1}{l}{\multirow{2}{*}{$q_{query}$}} & \multicolumn{1}{l}{\multirow{2}{*}{$f_{query}^{0}$}} & \multicolumn{4}{c}{mAP@0.5$\uparrow$}      \\ \cline{3-6} 
    \multicolumn{1}{l}{}                             & \multicolumn{1}{l}{}                                 & 20k iter & 40k iter & 80k iter & 160k iter \\ \hline
    \multicolumn{2}{c}{VoteNet\cite{chen2021scan2cap}}                                                      & -        & -        & -        & 32.21     \\
    \multicolumn{2}{c}{VoteNet$^*$}                                                                         & -        & -        & -        & 44.96     \\ \cline{1-2}
    $q_{seed}$                                       & $\mathbf{0}$                                         & 28.26    & 37.27    & 43.41    & 48.18     \\
    $q_{vq}$                                         & $\mathbf{0}$                                         & 24.73    & 33.21    & 41.56    & 49.15     \\
    $q_{vq}$                                         & $f_{vq}$                                             & \textbf{32.73}    & \textbf{39.55}    & \textbf{47.63}    & \textbf{52.13}     \\
    \bottomrule
    \end{tabular}
    }
\end{table}

\myparagraph{How does the Vote Query Improve 3DETR?}
For fair comparisons, we first train a VoteNet\cite{qi2019votenet} and a 3DETR-m\cite{misra2021-3detr} model with the same training strategy mentioned described in section \ref{subsec:datasets,metric,implementation} as our baseline.
Because of the longer and more advanced training strategy, our re-implemented VoteNet (VoteNet$^*$ in Table \ref{tab:ablation_detection}) performs significantly better than the basic version introduced in \cite{chen2021scan2cap}.
All comparisons are made on the ScanNet\cite{dai2017scannet} validation set.

As mentioned above, we formulate the object queries into $\left(p_{query}, f^{0}_{query}\right)$ so that the seed queries in 3DETR-m\cite{misra2021-3detr} and our proposed vote query could be written as $\left(p_{seed}, \mathbf{0}\right)$ and $\left(p_{vq}, f_{vq}\right)$ respectively.
We also introduce one more variant of vote query $\left(p_{vq}, \mathbf{0}\right)$ which only introduces 3D spatial bias.
One can see from Table \ref{tab:ablation_detection} that the introduction of 3D spatial bias itself to the query position $p_{vq}$ leads to an improvement in detection (\textcolor{mygreen}{+0.97}\% mAP@0.5).
However, it converges slower in the earlier training procedure than in the 3DETR-m baseline, inferring the vote query generation module is not well learned to predict accurate spatial offset estimations at early training epochs.
Besides, we can see a boost in both convergence and performance (\textcolor{mygreen}{+2.98}\% mAP@0.5) when we aggregate local contents for initial query feature $f_{vq}$ as well.
The overall performance of Vote2Cap-DETR is \textcolor{mygreen}{+3.95}\% mAP@0.5 higher than the 3DETR-m baseline, and \textcolor{mygreen}{+7.17}\% mAP@0.5 higher than the widely adopted VoteNet baseline.

\myparagraph{How does 3D Context Feature Help Captioning?}
Because the evaluation protocol of 3D dense captioning depends on both the localization and caption generation capability, we freeze all parameters other than the caption head and train with the standard cross entropy loss for a fair comparison.
Specifically, we employ the object-centric decoder\cite{wang2022spacap3d} as our baseline, which is a transformer-based model that can generate captions with an object's feature as the prefix of the caption.
In Table \ref{tab:ablation_caption_memory}, ``-'' refers to the object-centric decoder baseline, ``global'' naively involves all context tokens extracted from the scene encoder in the decoder, and ``local'' is our proposed \textbf{D}ual \textbf{C}lued \textbf{C}aptioner (DCC) that incorporates a vote query's $k_s$ ($k_s=128$ empirically) nearest context tokens extracted from the scene encoder.

Results show that the caption generation performance benefits from the introduction of additional contextual information.
Moreover, comparing to the naive inclusion of contextual information from the whole scene, the introduction of local context yields better results, which supports our motivation that considering the close surroundings of an object is crucial when describing it.

\begin{table}[htbp]
    \centering
    \caption{
        \textbf{Different keys for caption generation on the ScanRefer validation set.}
        We find out that introducing additional contextual information helps generate more informative captions.
        Since 3D dense captioning is more object-centric, introducing vote queries' local contextual feature is a better choice.
    }
    \label{tab:ablation_caption_memory}
    \resizebox{\linewidth}{!}{
    \begin{tabular}{ccccccccccc}
    \toprule
    \multirow{2}{*}{key} &  & \multicolumn{4}{c}{IoU=0.25}                                      &  & \multicolumn{4}{c}{IoU=0.5}                                       \\ \cline{3-6} \cline{8-11} 
                         &  & C$\uparrow$    & B-4$\uparrow$  & M$\uparrow$    & R$\uparrow$    &  & C$\uparrow$    & B-4$\uparrow$  & M$\uparrow$    & R$\uparrow$    \\ \cline{1-1} \cline{3-6} \cline{8-11} 
    -                    &  & 68.62          & 38.61          & 27.67          & 58.47          &  & 60.15          & 34.02          & 25.80          & 53.82          \\
    global               &  & 70.05          & 39.23          & 27.84          & 58.44          &  & 61.20          & 34.66          & 25.93          & 53.79          \\
    local                &  & \textbf{70.42} & \textbf{39.98} & \textbf{27.99} & \textbf{58.89} &  & \textbf{61.39} & \textbf{35.24} & \textbf{26.02} & \textbf{54.12} \\ \bottomrule
    \end{tabular}
    }
\end{table}

\begin{figure}[htbp]
	\centering
	\includegraphics[width=\linewidth]{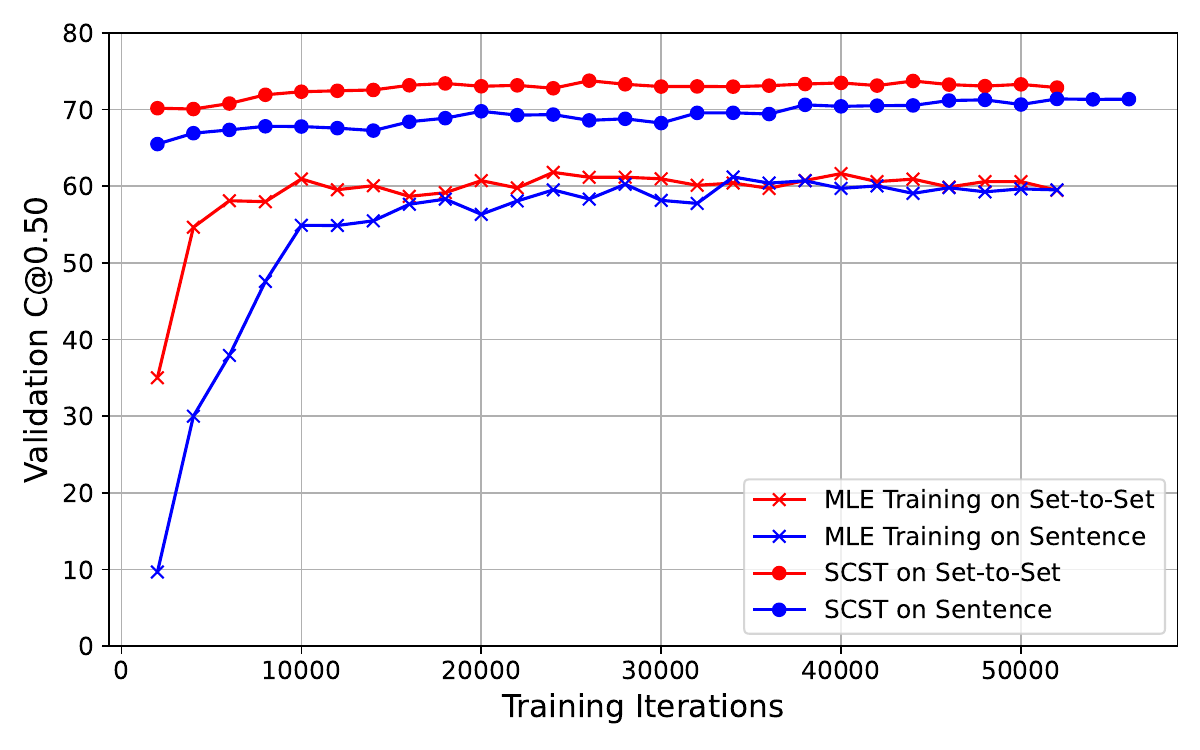}
	\caption{
	\textbf{Set-to-Set training and model convergence on the ScanRefer validation set.} 
	We analyze the convergence of two different training strategies with MLE training and SCST.
	Set-to-Set training enables a larger batch size for captioning and accelerates convergence.
	}
	\label{fig:set-to-set}
\end{figure}

\myparagraph{Does Set-to-Set Training Benefit 3D Dense Captioning?}
To analyze the effectiveness of set-to-set training, we use a smaller learning rate ($10^{-6}$) for all parameters other than the caption head and freeze these parameters during SCST.
In Table \ref{tab:set-to-set-training} We refer to the conventional training strategy widely used in previous studies \cite{chen2021scan2cap,wang2022spacap3d} as ``Sentence Training'', which traverses through all sentence annotations in the dataset.
As shown in Figure \ref{fig:set-to-set}, our proposed ``Set-to-Set'' training achieves comparable results with the traditional strategy during MLE training and converges faster because of a larger batch size on the caption head.
This also contributes to SCST.

\begin{table}[htbp]
    \centering
    \caption{
    \textbf{Set to Set training and performance on the ScanRefer validation set.} 
    We compare our proposed set-to-set training with traditional ``Sentence Training'', which traverses through all sentence annotations.
    We achieve comparable performance with MLE training, and 2.38\% C@0.5 improvement with SCST.
    }
    \label{tab:set-to-set-training}
    \resizebox{\linewidth}{!}{
    \begin{tabular}{cccccc}
    \toprule
    Training                 & $\mathcal{L}_{des}$   & C@0.5$\uparrow$ & B-4@0.5$\uparrow$ & M@0.5$\uparrow$ & R@0.5$\uparrow$ \\ \hline
    Sentence                 & \multirow{2}{*}{MLE}  & 61.21           & \textbf{35.35}    & 26.12           & \textbf{54.52}  \\
    Set-to-Set               &                       & \textbf{61.81}  & 34.46             & \textbf{26.22}  & 54.40           \\ \hline
    Sentence                 & \multirow{2}{*}{SCST} & 71.39           & 37.57             & 26.01           & 54.28           \\
    Set-to-Set               &                       & \textbf{73.77}  & \textbf{38.21}           & \textbf{26.64}        & \textbf{54.71}  \\ \bottomrule
    \end{tabular}
    }
\end{table}

\myparagraph{End-to-End Training from Scratch.}
Vote2Cap-DETR also enables end-to-end training from scratch for 3D dense captioning. 
However, both ScanRefer and Nr3D are annotated on limited scenes (562/511 scenes) for training; thus, directly training Vote2Cap-DETR from scratch will underperform given to satisfy two objectives simultaneously.
As experiments shown on ScanRefer in Table \ref{tab:end-to-end}, the greedy strategy we choose by pre-training the backbone on the detection task serves as a good pre-requisite for captioning to achieve better performance.
\begin{table}[htbp]
    \centering
    \caption{
    \textbf{Ablation study for training strategies on the ScanRefer validation set.}
    The greedy strategy we choose by pre-training the detection head as a good pre-requisite for captioning, achieves better performance than directly end-to-end training from scratch.
    }
    \label{tab:end-to-end}
    \resizebox{\linewidth}{!}{%
    \begin{tabular}{cccccccc}
    \toprule
    pretrain/end2end  & C@0.5$\uparrow$ & B-4@0.5$\uparrow$ & M@0.5$\uparrow$ & R@0.5$\uparrow$ & AP@0.5$\uparrow$ & AR@0.5$\uparrow$ \\ \toprule
    end2end     & 52.15      & 28.87        & 24.68      & 49.76      & 46.68     & 62.17     \\
    pretrain+end2end   & \textbf{62.03}      & \textbf{34.90}        & \textbf{26.06}      & \textbf{54.33}      & \textbf{51.26}     & \textbf{67.57}      \\
    \bottomrule
    \end{tabular}%
    }
\end{table}

\myparagraph{Is Vote2Cap-DETR Robust to NMS?}
Similar to other DETR works, the set loss encourages the model to produce sparse and non-duplicate predictions.
In Table \ref{tab:effect-nms}, we present a comparison on evaluating both 3D dense captioning (C@0.5) and detection (mAP50, AR50).
Since the $m@kIoU$ metric (Eq. \ref{eq:m@kIoU}) does not contain any penalties on redundant predictions, getting rid of NMS\cite{neubeck2006nms} results in performance growth in C@0.5.
Results demonstrate that Vote2Cap-DETR exhibits higher stability compared with VoteNet-based 3D dense captioning methods (\textit{i.e.} SpaCap3D\cite{wang2022spacap3d}, and 3DJCG\cite{cai20223djcg}) without the presence of NMS.
\begin{table}[htbp]
    \centering
    \caption{
    \textbf{Effect of NMS.} We analyze whether the absence of NMS affects the 3D dense captioning (C@0.5) as well as the model's detection performance (mAP50, AR50) on the ScanRefer validation set.
    }
    \label{tab:effect-nms}
    \resizebox{\linewidth}{!}{
    \begin{tabular}{ccccccccc}
    \toprule
    \multirow{2}{*}{Models} & \multicolumn{3}{c}{w/ NMS}                              & \multicolumn{3}{c}{w/o NMS}                            \\ \cline{2-7} 
                            & C@0.5$\uparrow$ & mAP50$\uparrow$ & AR50$\uparrow$ & C@0.5$\uparrow$ & mAP50$\uparrow$ & AR50$\uparrow$ \\ \hline
    SpaCap3D                & 43.93           & 37.77             & 53.96            & 51.35           & 23.30             & 64.14            \\
    3DJCG                   & 50.22           & 47.58             & 62.12            & 54.94           & 30.03             & \textbf{68.69}            \\
    Vote2Cap-DETR           & \textbf{70.63}           & \textbf{52.79}             & \textbf{66.09}            & \textbf{71.57}           & \textbf{52.82}             & 67.80            \\ \bottomrule
    \end{tabular}
    }
\end{table}

\subsection{Ablation Studies on Vote2Cap-DETR++}
\label{subsec:ablations-vote2cap++}

\begin{figure*}[htbp]
    \centering
    \hfill
    \begin{minipage}[t]{0.24\linewidth}
        \centering
        \includegraphics[width=\linewidth]{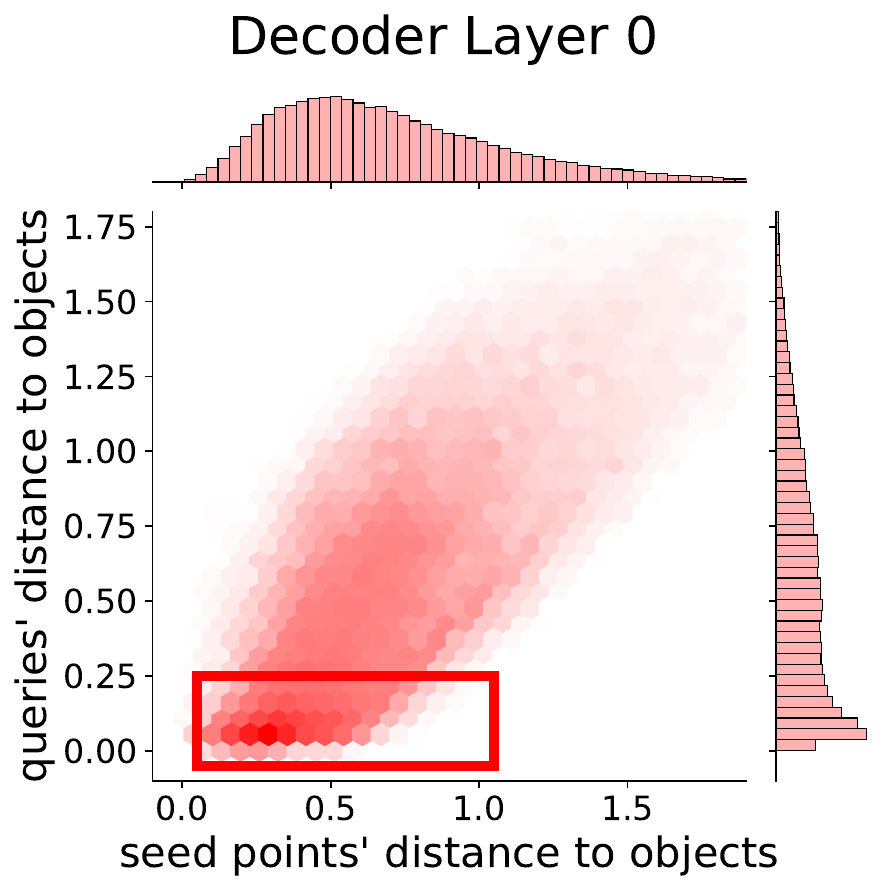}
    \end{minipage}
    \hfill
    \begin{minipage}[t]{0.24\linewidth}
        \centering
        \includegraphics[width=\linewidth]{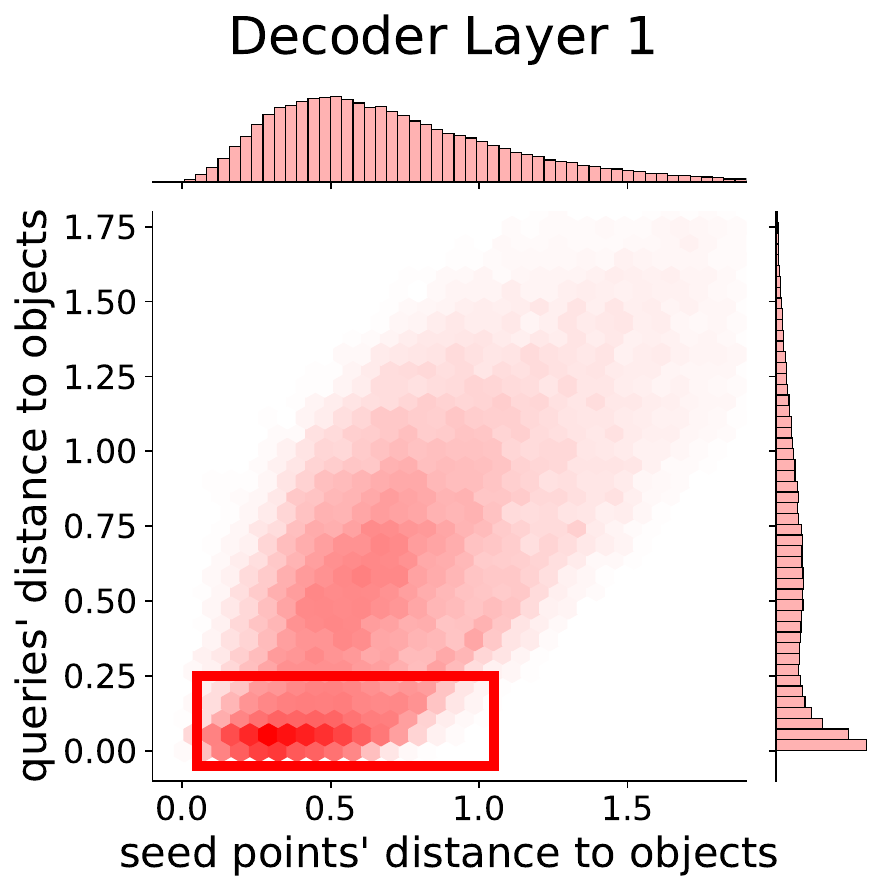}
    \end{minipage}
    \hfill
    \begin{minipage}[t]{0.24\linewidth}
        \centering
        \includegraphics[width=\linewidth]{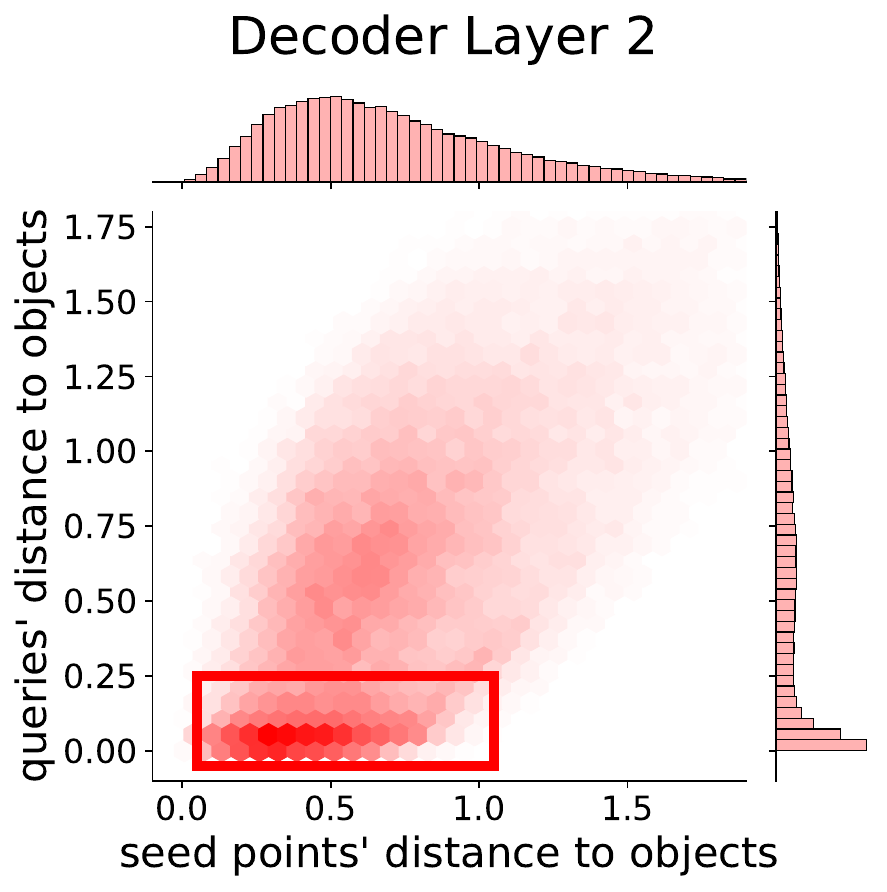}
    \end{minipage}
    \hfill
    \begin{minipage}[t]{0.24\linewidth}
        \centering
        \includegraphics[width=\linewidth]{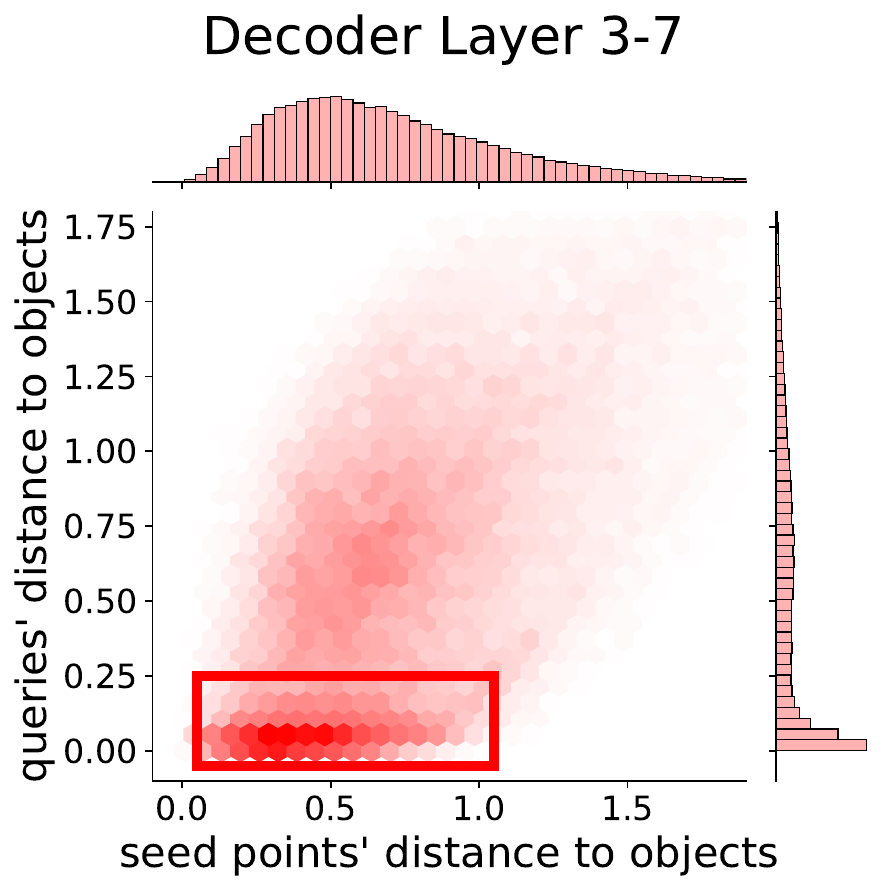}
    \end{minipage}
    \caption{
        \textbf{The spatial distribution of vote queries in different decoder layers.}
        For each seed point, we find its nearest object, and calculate its corresponding vote query's spatial distance to that object.
        One can see that as a layer goes deeper in the decoder, more and more vote queries are getting close to the objects.
	}
    \label{fig:query decoder layer}
\end{figure*}

\begin{figure*}[htbp]
	\centering
	\includegraphics[width=\linewidth]{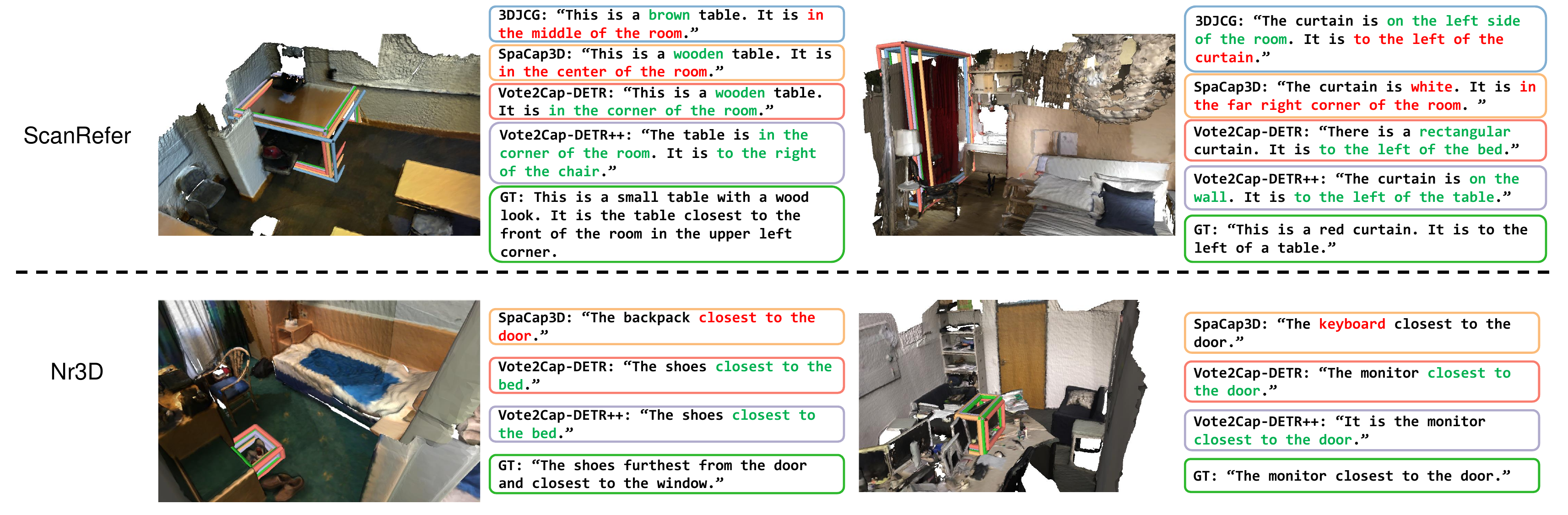}
	\caption{
    	\textbf{Qualitative results of our proposed methods on ScanRefer and Nr3D validation set.}
        We showcase the localization results with the corresponding captions generated.
        One can see that our proposed methods are able to generate tight bounding boxes close to the object surfaces and accurate descriptions.
    }
	\label{fig:qualitative}
\end{figure*}

\whatsnew{
In this section, we also provide thorough experiments under the same setting as section \ref{subsec:ablations-vote2cap} to study different components proposed in Vote2Cap-DETR++.
}

\begin{table}[htbp]
    \caption{
        \textbf{Per layer performance in Vote2Cap-DETR on the ScanNet validation set.}
        We evaluate the performance of different decoder layers.
        There is a significant performance growth in the first three layers, while the latter layers perform somewhat similarly.
    }
    \label{tab:per-layer}
    \centering
    \resizebox{\linewidth}{!}{
    \begin{tabular}{ccccccccc}
    \toprule
    Layer-id          & 0     & 1     & 2     & 3     & 4     & 5     & 6     & 7     \\
    \hline
    mAP@0.5$\uparrow$ & 48.17 & 49.91 & 51.20 & 52.11 & 52.50 & 52.26 & 52.31 & 52.49 \\
    AR@0.5$\uparrow$  & 67.72 & 67.99 & 67.80 & 68.76 & 68.89 & 69.06 & 69.06 & 69.30 \\
    \bottomrule
    \end{tabular}
    }
\end{table}

\myparagraph{Which Layer Shall We Refine the Queries?}
\whatsnew{
To better analyze the effectiveness of the iterative spatial refinement strategy for vote queries in Vote2Cap-DETR++, we first evaluate a per-layer detection performance of different decoder layers in Vote2Cap-DETR on the ScanNet\cite{dai2017scannet} validation set in Table \ref{tab:per-layer}.
One can see that the performance grows as the layer goes deeper.
Concurrently, the performance of the first three decoder layers is relatively poor and varies greatly, while the performance of the last five layers are similar ($\ge$ 52.0\% mAP@0.5).
Thus, we compare different combinations in Table \ref{tab:revoting}.
Here, the baseline model marked ``-'' does not perform any refinement step, which downgrades to Vote2Cap-DETR.
The model marked ``\textit{all}'' implies that we adopt the refinement strategy along all the decoder layers.
The results marked $[0, 1]$ and $[0, 1, 2]$ stand for refining vote queries in the first two (layer 0 and 1), and the first three layers (layer 0, 1, and 2) respectively.
Experiments show that adopting the spatial refinement in the first three decoder layers achieves the best performance.
}

\begin{table}[htbp]
    \centering
    \caption{
        \textbf{Analysis of different strategies for iterative spatial refinement on vote queries.}
        We apply the refinement strategy to update the spatial localization of vote queries in different decoder layers.
        Experiments show that refining the spatial location of vote queries in the first three decoder layers leads to better performance.
    }
    \label{tab:revoting}
    \resizebox{0.9\linewidth}{!}{
        \begin{tabular}{lccccc}
        \toprule
                \multicolumn{1}{c}{\multirow{2}{*}{Refine Layer}} & \multicolumn{2}{c}{IoU=0.25} & & \multicolumn{2}{c}{IoU=0.5}  \\ \cline{2-3} \cline{5-6}
        \multicolumn{1}{c}{}                                & mAP$\uparrow$ & AR$\uparrow$ &     & mAP$\uparrow$ & AR$\uparrow$ \\ \hline
        -                                                   & 69.61         & \textbf{87.20}&     & 52.13         & 69.12        \\
        $[0, 1]$                                            & 69.59         & 85.43        &     & 54.59         & 70.15        \\
        $[0, 1, 2]$                                         & \textbf{70.52}         & 85.64        &     &\textbf{55.48}         & \textbf{70.89}        \\
        \textit{all}                                        & 69.96         & 85.63        &     & 54.55         & 69.81        \\
        \bottomrule
        \end{tabular}
    }
    \vspace{-6mm}
\end{table}

\myparagraph{Per-Layer Detection Comparison between Vote2Cap-DETR and Vote2Cap-DETR++.}
\whatsnew{
We compare the detection performance among different decoder layers in Vote2Cap-DETR and Vote2Cap-DETR++ in Table \ref{tab:per-layer-comparison}.
The first decoder layer has similar performance, while the following three layers of Vote2Cap-DETR++ perform far better than those in Vote2Cap-DETR (\textcolor{mygreen}{+1.66}\%, \textcolor{mygreen}{+2.46}\%, \textcolor{mygreen}{+2.23}\% mAP@0.5).
This further indicates that the first three layers efficiently move the queries spatially to precise locations close to objects, leading to box estimations of higher quality.
}
\begin{table*}[htbp]
    \caption{
        \textbf{Decoders' per-layer Detection Performance Comparison between Vote2Cap-DETR and Vote2Cap-DETR++ on ScanNet Validation Set.}
        The first decoder layer has similar performance between the two models, while the performance of the latter three decoder layers have a bigger difference.
        This indicates that the refinement strategy we adopted on the spatial location of queries in the first three decoder layers encourages high-quality box estimations.
    }
    \label{tab:per-layer-comparison}
    \centering
    \resizebox{0.8\linewidth}{!}{
    \begin{tabular}{cccccccccc}
    \toprule
    \multicolumn{2}{c}{Layer-id}                         & 0     & 1     & 2     & 3     & 4     & 5     & 6     & 7     \\ \hline
    \multirow{3}{*}{mAP@0.5$\uparrow$} & Vote2Cap-DETR   & 48.17 & 49.91 & 51.20 & 52.11 & 52.50 & 52.26 & 52.31 & 52.49 \\
                                       & Vote2Cap-DETR++ & 48.07 & 51.57 & 53.66 & 54.34 & 55.11 & 55.34 & 55.52 & 55.48 \\ 
                                       & $\Delta$ & \textcolor{myred}{-0.10\%} & \textcolor{mygreen}{+1.66\%} & \textcolor{mygreen}{+2.46\%} & \textcolor{mygreen}{+2.23\%} & \textcolor{mygreen}{+2.61\%} & \textcolor{mygreen}{+3.12\%} & \textcolor{mygreen}{+3.21\%} & \textcolor{mygreen}{+2.99\%} \\ 
    \bottomrule
    \end{tabular}
    }
\end{table*}

\myparagraph{Spatial Location of Queries in Different Decoder Layers.}
\whatsnew{
We showcase the distribution of spatial locations for vote queries in different decoder layers. 
As mentioned above, the last five decoder layers share the same spatial location of vote queries.
For each seed point, we find its nearest object, and calculate its corresponding vote query's spatial distance to that object in Figure \ref{fig:query decoder layer}.
Results show that the seed points that are initially far away from the objects get closer and closer to nearby objects as the decoder layer goes deeper, resulting in higher-quality box estimations with faster convergence.
}

\myparagraph{Comparing with Other 3DETR Attempts.}
\whatsnew{
Since there are few works that directly improve 3DETR\cite{misra2021-3detr}, we compare our proposed Vote2Cap-DETR and Vote2Cap-DETR++ with the hybrid matching strategy\cite{jia2022hybriddetrs} and the learnable anchor points\cite{wang2022anchor} in Table \ref{tab:other 3detr attempts}.
In practice, the hybrid matching strategy maintains another set of object queries that are supervised by one-to-many label assignment, while the learnable anchor points are randomly initialized following \cite{wang2022anchor}.
As shown in Table \ref{tab:other 3detr attempts}, both methods are inferior to either of our proposed methods.
Though the hybrid matching accelerates the training of 3DETR-m in the early training epochs, it still falls behind Vote2Cap-DETR when it converges.
Further, the advanced version, Vote2Cap-DETR++, has a faster convergence speed in the early stage than any other methods, and even a better detection performance (\textcolor{mygreen}{+3.35}\% mAP@0.5) than Vote2Cap-DETR when the model converges.
}
\begin{table}[htbp]
    \centering
    \caption{
        \textbf{Comparison with other 3DETR attempts on the ScanNet validation set.}
        We compare the detection performance of our proposed methods with different modifications that improve 3DETR at the 20k, 40k, 80k, 160k -\textit{th} iteration.
    }
    \label{tab:other 3detr attempts}
    \resizebox{\linewidth}{!}{%
    \begin{tabular}{cccccc}
    \toprule
    \multirow{2}{*}{Model}   & \multirow{2}{*}{Modification} & \multicolumn{4}{c}{mAP@0.5$\uparrow$}      \\ \cline{3-6} 
                             &                               & 20k iter & 40k iter & 80k iter & 160k iter \\ \hline
    \multirow{3}{*}{3DETR-m} & -                             & 28.26    & 37.27    & 43.41    & 48.18     \\
                             & hybrid\cite{jia2022hybriddetrs}                        & 35.10    & 42.72    & 45.83    & 47.50     \\
                             & anchor\cite{wang2022anchor}                        & 22.94    & 28.85    & 35.44    & 40.06     \\ \cline{1-1}
    \multicolumn{1}{l}{\textbf{\textit{Ours:}}}    \\
    Vote2Cap-DETR            & -                             & 32.73    & 39.55    & 47.63    & 52.13     \\
    Vote2Cap-DETR++          & -                             & \textbf{40.09}    & \textbf{44.50}    & \textbf{50.09}    & \textbf{55.48}    \\ 
        \bottomrule
    \end{tabular}
    }
\end{table}

\myparagraph{Design of Decoupling Queries.}
\whatsnew{
We conduct studies on different designs for task-specific queries on the ScanRefer validation set in Table \ref{tab:decouple query}.
The first row refers to our baseline method that generates captions and localizes objects with shared queries as Vote2Cap-DETR.
One can see that standalone decoupling of the queries leads to a performance drop.
However, when we link the [CAP] queries with [LOC] queries through token-wise projection, we witness a relative performance improvement of \textcolor{mygreen}{+1.57}\% C@0.5.
}
\begin{table}[htbp]
    \centering
    \caption{
        \textbf{Effectiveness of decoupled-and-correspond query designs.}
        Without linking the [CAP] queries with [LOC] queries through token-wise projection, standalone decoupling the queries will lead to a performance drop.
    }
    \label{tab:decouple query}
    \resizebox{\linewidth}{!}{
    \begin{tabular}{ccccccc}
    \toprule
    decouple     & correspond   & C@0.5$\uparrow$ & B-4@0.5$\uparrow$ & M@0.5$\uparrow$ & R@0.5$\uparrow$ & mAP@0.5$\uparrow$\\ \hline
    -            & -            & 66.01           & \textbf{37.41}    & 26.62           & 55.33           & 58.18            \\
    $\checkmark$ & -            & 65.38           & 37.07             & 26.71           & 55.41           & 58.67            \\
    $\checkmark$ & $\checkmark$ & \textbf{67.58}  & 37.05             & \textbf{26.89}  & \textbf{55.64}  & \textbf{58.83}            \\
    \bottomrule
    \end{tabular}
}
\end{table}

\myparagraph{How does additional 3D Spatial Information Help Captioning?}
\whatsnew{
To address the effectiveness of different designs of the spatial information injection, we evaluate the performance of different strategies on ScanRefer\cite{chen2020scanrefer} validation set in Table \ref{tab:captioner spatial} with a frozen backbone.
In Table \ref{tab:captioner spatial}, the model in the first row downgrades to DCC since there is no additional spatial information injected into the model.
It can be seen that introducing an additional position embedding token $\mathcal{V}^{q}_{pos}$ as the caption prefix largely improves the quality of the generated captions.
We also find out that compared with absolute 3D positional encoding, using a shared ranking-based position embedding of context tokens $\mathcal{V}^{s}_{pos}$ further improves the captioning performance.
}
\begin{table}[htbp]
    \centering
    \caption{
        \textbf{The Effectiveness of Spatial Information on Caption Generation.}
        We evaluate different models on the ScanRefer validation set.
        Both the introduction of a vote query's absolute position, and the ranking-based contextual position lead to a better quality of captions.
    }
    \label{tab:captioner spatial}
    \resizebox{\linewidth}{!}{
    \begin{tabular}{cccccc}
    \toprule
    \multicolumn{2}{c}{Modification}                                                          & \multirow{2}{*}{C@0.5$\uparrow$} & \multirow{2}{*}{B-4@0.5$\uparrow$} & \multirow{2}{*}{M@0.5$\uparrow$} & \multirow{2}{*}{R@0.5$\uparrow$} \\ \cline{1-2}
    \multicolumn{1}{l}{$\mathcal{V}^{q}_{pos}$} & \multicolumn{1}{l}{$\mathcal{V}^{s}_{pos}$} &                                  &                                    &                                  &                                  \\ \hline
    -                                           & -                                           & 62.15                            & 35.61                              & 26.31                            & 54.83                            \\
    $\checkmark$                                &  -                                          & 64.34                            & 36.68                              & 26.37                            & 55.18                            \\
    $\checkmark$                                & \textit{abs.}                               & 61.86                   & 36.06                     & 26.18                   & 54.95                   \\
    $\checkmark$                                & \textit{rank}                               & \textbf{65.00}                   & \textbf{36.97}                     & \textbf{26.45}                   & \textbf{55.29}                   \\
    \bottomrule
    \end{tabular}
    }
\end{table}

\myparagraph{Per-class Detection Performance.}
We list per-class detection AP results of the re-implemented VoteNet\cite{qi2019votenet}, 3DETR-m\cite{misra2021-3detr}, and our proposed Vote2Cap-DETR and Vote2Cap-DETR++ on ScanNet\cite{dai2017scannet} validation set under an IoU threshold of 0.5 in Table \ref{tab:ScanNet AP per class}.
The overall performance is listed in Table \ref{tab:ablation_detection}.

\begin{table*}[htbp]
    \centering
    \caption{\textbf{Per-class AP under IoU threshold of 0.5 on ScanNet validation scenes.}}
    \label{tab:ScanNet AP per class}
    \resizebox{\linewidth}{!}{
    \begin{tabular}{ccccccccccccccccccc}
    \toprule
    Method        & cabinet & bed   & chair & sofa  & table & door  & window & bookshelf & picture & counter & desk  & curtain & refrigerator & shower curtain & toilet & sink  & bathtub & others \\ \hline
    VoteNet\cite{qi2019votenet}       & 21.41   & 78.41 & 78.47 & 74.44 & 55.42 & 34.68 & 14.91  & 29.80     & 9.04    & 16.57   & 51.12 & 34.62   & 40.12        & 45.82          & 89.93  & 37.23 & 83.41   & 13.79  \\
    3DETR\cite{misra2021-3detr}         & 26.30   & 75.78 & 82.19 & 59.15 & 62.25 & 39.16 & 21.47  & 33.14     & 16.45   & 34.41   & 49.68 & 38.34   & 42.83        & 33.33          & 88.68  & 52.62 & 82.41   & 29.06  \\ \cline{1-1}
    \multicolumn{1}{l}{\textbf{\textit{Ours:}}} \\
    Vote2Cap-DETR & 31.98   & 81.48 & 85.80 & 64.37 & 65.20 & 41.19 & 28.47  & 39.81     & 22.94   & 39.02   & 54.46 & 36.66   & 40.19        & 56.10          & 87.97  & 44.38 & 85.12   & 33.28  \\
    Vote2Cap-DETR++ & 34.10   & 79.81 & 85.96 & 77.00 & 65.94 & 46.22 & 36.38  & 42.93     & 22.14   & 36.20   & 53.33 & 46.29   & 44.34        & 63.17          & 89.66  & 53.78 & 89.29   & 32.16  \\
    \bottomrule
    \end{tabular}
    }
\end{table*}

\subsection{Qualitative Results}

\label{subsec:viz}

In this section, we mainly provide some qualitative results to visualize the effectiveness of our proposed methods.

\begin{figure*}[htbp]
	\centering
	\includegraphics[width=\linewidth]{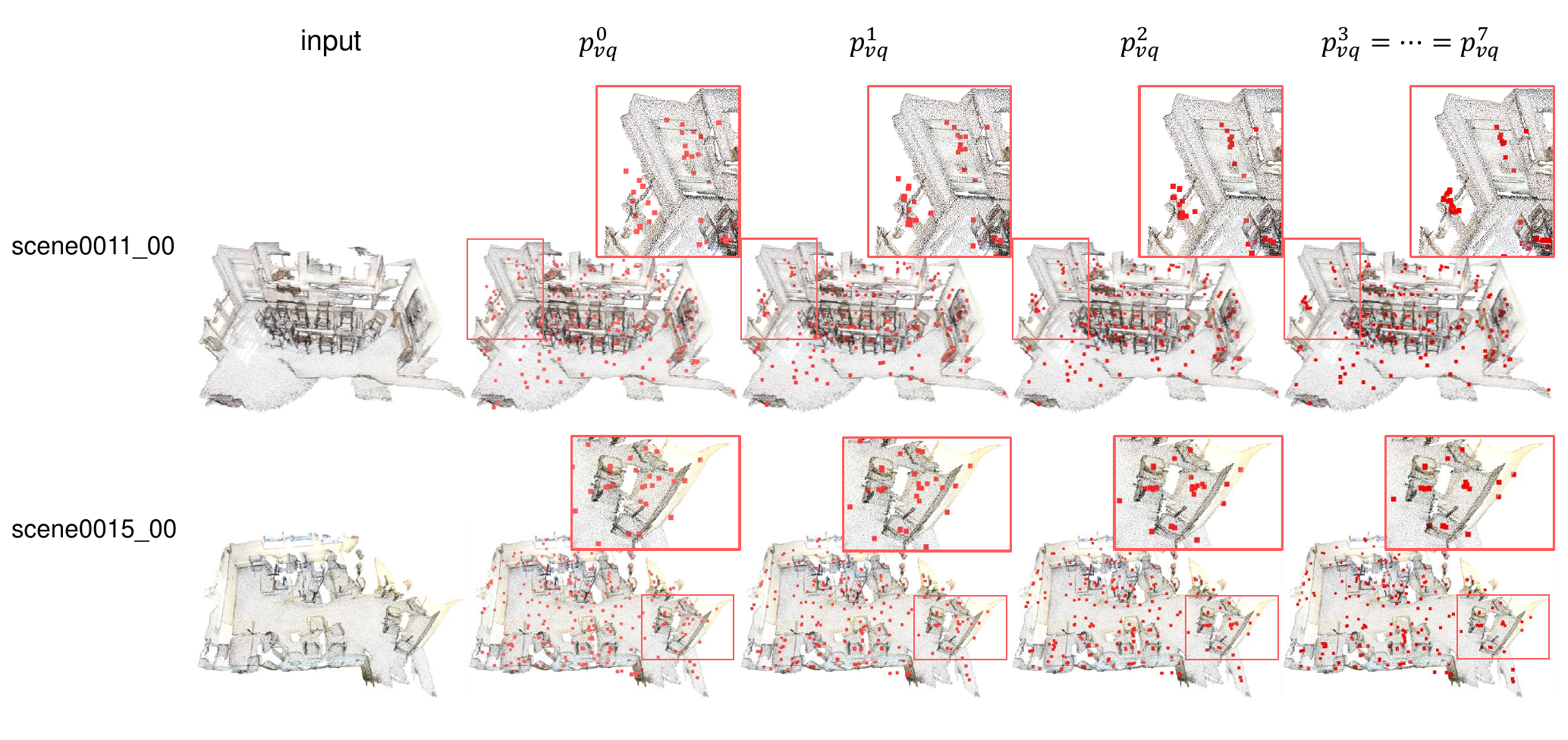}
	\caption{
    	\textbf{Visualization of the queries' spatial locations in different decoder layers.}
        We visualize the input point clouds as well as the spatial locations of queries in each decoder layer.
        As the decoder goes deeper, the queries get more concentrated to the object centers.
    }
	\label{fig:viz-vote-refinement}
\end{figure*}

\begin{figure*}[htbp]
	\centering
	\includegraphics[width=\linewidth]{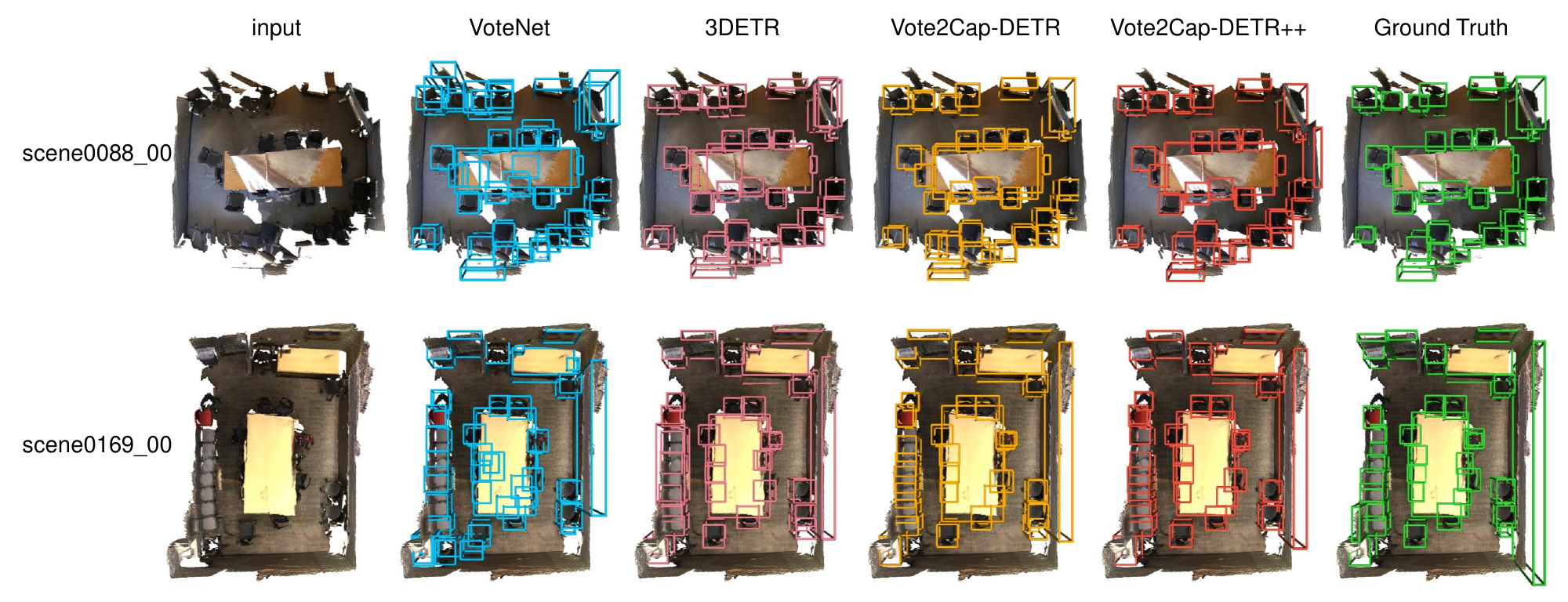}
	\caption{
    	\textbf{Visualization of box estimations.}
        We showcase the localization results for different input scenes.
        Our proposed methods, Vote2Cap-DETR and Vote2Cap-DETR++, can generate tight bounding boxes close to the ground truth.
    }
	\label{fig:viz-localization}
\end{figure*}

\myparagraph{Qualitative Results on ScanRefer and Nr3D.}
We showcase several localization results and the captions generated in Figure \ref{fig:qualitative}.
One can see that our proposed methods are able to generate tight bounding boxes close to the object surfaces and accurate descriptions.

\myparagraph{Visualization for the Queries' Spatial Locations.}
\whatsnew{
We have visualized the spatial location of vote queries in different decoder layers in Figure \ref{fig:viz-vote-refinement}.
The deeper the decoder layer is, the closer the vote queries are to the box centers.
}

\myparagraph{Visualization of Object Localization Results.}
We also showcase several object localization results of different methods in Figure \ref{fig:viz-localization}.
Both of our proposed methods, Vote2Cap-DETR and Vote2Cap-DETR++, are able to generate tight bounding boxes close to the ground truth.

\section{Limitations and Open Questions}
\label{sec:limitations}
Though we have proposed two effective non-``detect-then-describe'' methods for 3D dense captioning, the captions do not have much diversity because of the limited text annotations, beam search, and self-critical sequence training with the CiDEr reward.
We believe that multi-modal pre-training on 3D vision-language tasks with more training data and the utilization of \textbf{L}arge \textbf{L}anguage \textbf{M}odels(LLM) trained on large corpus would increase the diversity of the generated captions.
Additionally, other reward functions designed for 3D dense captioning will increase the diversity among object descriptions in the same scene.
We will leave these topics for future study.

\section{Conclusions}
\label{sec:conclusion}
\whatsnew{
In this work, we decouple the caption generation from caption generation, and propose a set of two transformer-based approaches, namely Vote2Cap-DETR and Vote2Cap-DETR++, for 3D dense captioning.
Comparing with the sophisticated and explicit relation modules in conventional ``detect-then-describe'' pipelines, our proposed methods efficiently capture the object-object and object-scene relation through the attention mechanism.
The preliminary model, Vote2Cap-DETR, decouples the decoding process to generate captions and box estimations in parallel.
We also propose vote queries for fast convergence, and develop a novel lightweight query-driven caption head for informative caption generation.
In the advanced model, Vote2Cap-DETR++, we further decouple the queries to capture task-specific features for object localization and description generation.
Additionally, we introduce an iterative spatial refinement strategy for vote queries, and insert 3D spatial information for more accurate captions.
Extensive experiments on two widely used datasets validate that both the proposed methods surpass prior ``detect-then-describe'' pipelines by a large margin.
}

{\small
\bibliographystyle{plain}
\bibliography{reference}
}

\end{document}